%% file: main.tex
\documentclass[letterpaper, 10 pt, conference]{ieeeconf}  %

\IEEEoverridecommandlockouts                              %

\input{preamble}

\usepackage{xspace}
\usepackage[subtle,title]{savetrees}
\usepackage{soul}
\usepackage{booktabs}

\definecolor{lightblue}{rgb}{0.9,0.9,1}

\newcommand{\algname}{NLMap\xspace}
\newcommand{\link}[1]{\textcolor{magenta}{\href{#1}{#1}}}

\definecolor{lightgray}{gray}{0.9}
\usepackage{listings}
\title{ \LARGE \bf
Open-vocabulary Queryable Scene Representations for Real World Planning
\vspace*{5mm}}

\author{Boyuan Chen$^{1,\dagger}$, Fei Xia$^2$, Brian Ichter$^2$, Kanishka Rao$^2$, Keerthana Gopalakrishnan$^2$,\\ Michael S. Ryoo$^2$, Austin Stone$^2$, Daniel Kappler$^1$ \vspace*{-6mm}
\thanks{$^1$Everyday Robots, $^2$Robotics at Google, $^\dagger$ MIT. Emails: \url{boyuanc@mit.edu},\url{xiafei@google.com}}
\date{}
}

\begin{document}

\maketitle
 \thispagestyle{empty}
\pagestyle{empty}

\begin{abstract}
Large language models (LLMs) have unlocked new capabilities of task planning from human instructions. However, prior attempts to apply LLMs to real-world robotic tasks are limited by the lack of grounding in the surrounding scene. In this paper, we develop \algname, an open-vocabulary and queryable scene representation to address this problem. %
\algname serves as a framework to gather and integrate contextual information into LLM planners, allowing them to see and query available objects in the scene before generating a context-conditioned plan. \algname first establishes a natural language queryable scene representation with Visual Language models (VLMs). An LLM based object proposal module parses instructions and proposes involved objects to query the scene representation for object availability and location. An LLM planner then plans with such information about the scene. \algname allows robots to operate without a fixed list of objects nor executable options, enabling real robot operation unachievable by previous methods. Project website: \link{https://nlmap-saycan.github.io}

\end{abstract}

\section{Introduction}\label{sec:intro}

For robots to perform varied, real-world tasks, they must be able to comprehend diverse human commands and then act on these commands in the context of their environment.
Imagine a robot in a home environment tasked with ``water the plants in the living room''. It has to first identify relevant objects and locations within the scene (e.g., the watering can, the sink, and each potential plant) and then plan over these objects in sequential order (get the watering can, then go the sink, and then fill it up), conditioning on its affordances (e.g., can it carry a full watering can), and conditioning on the scene (e.g., how many plants there are, and where are they).
Semantic representation and downstream mobile manipulation planners capable of accessing this representation emerge as critical challenges in such a pipeline.

Semantic understanding is crucial for a robot to achieve long-horizon tasks in unstructured environments. %
Though a robot can avoid building a semantic representation by finding objects each time they are required, e.g., with Object Goal Navigation~\cite{chaplot2020object, anderson2018evaluation}, this repeated exploration can be inefficient. A persistent scene representation on the other hand avoids this exploration, but past works are generally limited to locating object categories known during the construction of the representation and may not encode the open-vocabulary objects that arise from human queries, such as in ``bring me the purple unicorn plush toy''. Recent progress in contrastively trained visual language models offers a promising solution to open-ended scene presentation. Contrastive Language-Image Pre-training (CLIP)~\cite{radford2021learning} models are trained on image-language associations and can provide open-vocabulary image understanding and object detection~\cite{gu2021open}. They have demonstrated impressive zero-shot classification performance and thus might be used to build a semantic representation in a zero-shot manner.

Another challenge lies in connecting the semantic scene representation to a planning algorithm that is capable of acting upon it. Recent progress in large language models (LLMs), has shown impressive few-shot performance in language comprehension, semantic understanding, and reasoning, as well as application to robotics problems like planning~\cite{huang2022language, saycan2022arxiv, huang2022inner} and instruction following~\cite{shah2022lm}.
Using such models in embodied settings can provide significant challenges, most critically because LLMs are not grounded in the physical world. For example, \cite{huang2022language} pioneers in using LLMs for planning, but it has no grounding in environmental context. In contrast, SayCan~\cite{saycan2022arxiv} showed how value functions of learned skills can provide such a grounding through selecting options scored highly by a language model and an affordance model. However, this is limited by the options provided and hardcoded knowledge of where objects exist.

\begin{figure}[t]
\setlength{\belowcaptionskip}{-12pt}
  \vspace{0.5em}
    \includegraphics[width=0.95\linewidth]{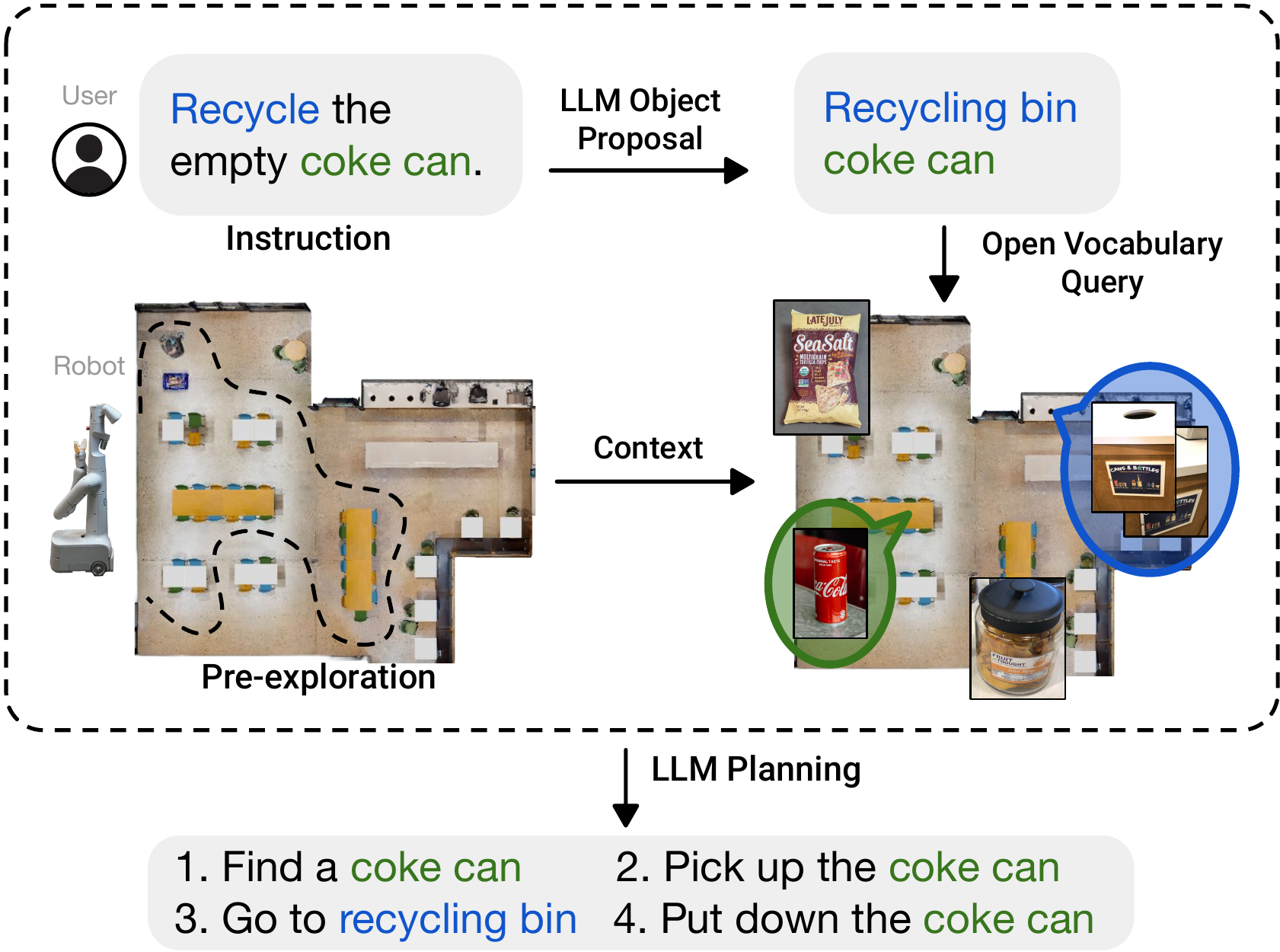}
    \caption{
        \footnotesize
        \textbf{\algname + SayCan overview.} We propose an open-vocabulary and queryable scene representation for real-world planning. A queryable scene representation is built from exploration. When the system receives a user query, it uses an LLM-based object proposal module to propose relevant objects to query the map. The returned object presence and location are used for LLM-based planning. We  benchmark the method on robots from \href{https://everydayrobots.com/}{Everyday Robots}.
    }
    \label{fig:teaser}
\end{figure}

\begin{figure*}[t]
\setlength{\belowcaptionskip}{-12pt}
  \vspace{0.5em}
  \begin{center}
    \includegraphics[width=0.95\textwidth]{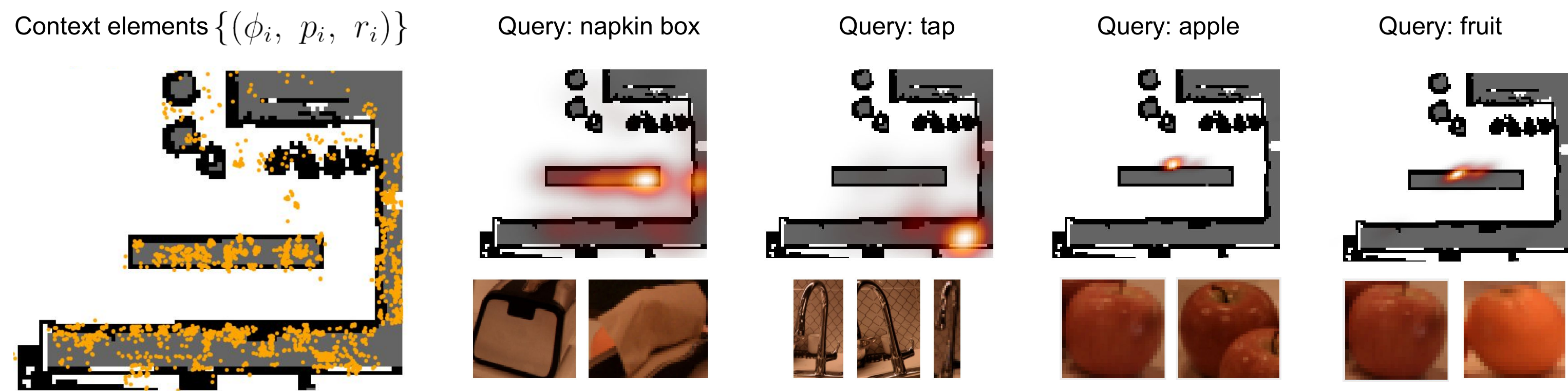}
    \end{center}
    \caption{
        \footnotesize
        \textbf{Natural Language Queryable Scene Representation}. The key design of \algname is to establish a queryable map. First, the agent explores the scene and provides a class-agnostic bounding box proposal based on objectness. We extract 512d CLIP features and 512d ViLD features of each bounding box and represent them as a feature point cloud $\mathcal{C}=\{(\phi_i, p_i, r_i)\}_{i=1...N}$. When queried with a piece of text, we visualize the heatmap of matches based on the alignment of text and visual features. Note that we can query with a single object name, or object families, such as ``snack" or ``fruit".
    \vspace{-2mm}\label{fig:map}
    }
\end{figure*}

In this work, we introduce Natural-Language Map (\algname), a flexible and language-queryable spatial semantic representation based on visual-language models including ViLD and CLIP and integrate with SayCan. 
We show that \algname grounds LLM-based planners in their environments, significantly improves long-horizon planning via natural language instructions in the open-world domain, and enables new tasks prior state-of-the-art algorithms failed to address.
To summarize, we make the following contributions:
\begin{enumerate}[wide, labelwidth=!, labelindent=0pt]
    \item We propose an open-vocabulary, queryable semantic representation based on ViLD and CLIP.
    \item We integrate \algname into a language-based planner to enable grounding on the context.
    \item We benchmark \algname + SayCan in a real-world kitchen, showing it is capable of performing $55$ tasks at $61.8\%$ success rate. Notably, $35$ of these tasks are impossible with previous state-of-the-art planners that do not have access to \algname.
\end{enumerate}

\section{Related Work}\label{sec:related}

\noindent\textbf{Semantic Scene Representations.}
Scene representation is a central theme in robot perception and planning. Semantic SLAM~\cite{civera2011towards, zhang2018semantic, bowman2017probabilistic} is an augmentation over traditional SLAM, it assigns semantic features over geometric features provided by SLAM (points, lines, planes). Many representations are proposed, ranging from a faithful 3D recontruction~\cite{xiazamirhe2018gibsonenv} of the environment, to more object-centric ones~\cite{runz2018maskfusion, mccormac2018fusion++}, such as object detection bounding boxes~\cite{qi2020imvotenet} and 3D bounding boxes~\cite{xu2018pointfusion}. Recently, topological maps~\cite{chen2021topological, chen2019behavioral} and scene graphs~\cite{wu2021scenegraphfusion, armeni20193d} emerge as an effective discrete representation of scenes.

One issue with those representations is that they cannot be queried with natural language. Interfacing with those scene representations requires reducing the object set to a closed set, indicating that they are not as useful for LLM-based planners and that they are limited in an open-vocabulary setting. In contrast, our work allows \emph{the scene representation to be queried at test time with natural language}. Concurrent work VLMaps~\cite{huang2022visual} also explores this concept, by fusing  pretrained visual-language model features into a geometric reconstruction of the scene. The representation is then used for visual-language navigation tasks via program synthesis.

\noindent\textbf{Object Goal Navigation.}
There is also a significant body of related work on object navigation, which focuses on flexible exploration to find objects in unknown scenes. A few of these algorithms construct a semantic map of the current region before planning in that region \cite{chaplot2020learning, chaplot2020neural, scenememorytransformer, chaplot2020object}. Map-based methods are modular and interpretable and hence easier to deploy in the real world. Other algorithms \cite{talukder2003real,santos1995visual,mousavian2019visual,chen2018learning, ramakrishnan2022poni, objectnavgoog} do not require a map and can decide where to go based directly on the current observations and memories, without maintaining a global representation of the environment. Recently, methods that leverage pre-trained image-text models can do zero-shot Object Goal Navigation~\cite{majumdar2022zson, gadre2022clip}. CoW~\cite{gadre2022clip} performs zero-shot object goal navigation by leveraging CLIP. LM-Nav~\cite{shah2022lm} uses three pretrained model to perform visual language navigation. Our work differs from Object Goal Navigation since the eventual goal is not purely finding objects, but using object presence and location information for planning. Our work can use the representation from a single exploration for many downstream planning tasks without the need to run Object Goal Navigation every time.

\noindent\textbf{Planning with Scene Representations.}
In task and motion planning, scene representations are often composed of predicates compatible with symbolic planners~\cite{garrett2015ffrob, garrett2020pddlstream}. Recent progresses attempt to build a symbolic and geometric scene graph to facilitate task and motion planning~\cite{zhu2021hierarchical}. However, they still require defining the objects in the scene. Recently LLM-based planners are more flexible~\cite{saycan2022arxiv, zeng2022socratic, huang2022inner} and do not require handcrafting predicates, however, they do not handle the complexity of open-vocabulary object proposal and require defining a set of objects involved in planning. They also fail to integrate perception in real robot experiments due to the difficulty of connecting unstructured natural language instruction to perception algorithms that need structured inputs.

\section{Problem Statement}\label{sec:problem}
In this work, we aim to efficiently fulfill high-level, natural-language instructions, such as ``Bring me a snack" or ``I spilled my coffee, can you help?''. This requires a robotic system to solve problems at the intersection of natural language comprehension, scene understanding, task planning, navigation, and manipulation. 
Recent work, SayCan~\cite{saycan2022arxiv}, has shown how large language models can be applied to such problems through world-grounding affordance functions, allowing LLMs to understand what a robot can do from a \emph{state}.
However, SayCan did not provide scene-scale affordance grounding, and thus cannot reason over what a robot can do in a \emph{scene}.
To that end, we address two core problems (i) how to maintain open-vocabulary scene representations that are capable of locating arbitrary objects and (ii) how to merge such representations within long-horizon LLM planners to imbue them with scene understanding.

\section{\algname + SayCan}\label{sec:alg}
We provide a high-level description of our algorithm in Listing~\ref{lst:algobox}. The design of each component is described below: 
\subsection{Scene Representation}\label{sec:alg_scene}
The scene representation is generated from an exploration phase of the unstructured scene, which our approach is agnostic to, but could be for example frontier exploration~\cite{yamauchi1998frontier} or pre-determined waypoints. During this exploration, \algname runs a class agnostic region proposal network as in ViLD~\cite{gu2021open} on all the observed RGB images. For each proposed region of interest (ROI) $I_i\in I_{1...N}$, our method uses an ensemble of VLM image encoders $\Phi_{1...M}$~\cite{gu2021open, radford2021learning} to extract image embeddings $\phi_i=[\Phi_j(I_i) |\ j\in 1...M]$. As shown in Fig.~\ref{fig:map}, such embedding can be queried with text at plan time since VLMs are capable of estimating the correlation between texts and images. 
In our setup we leverage CLIP~\cite{radford2021learning} and ViLD~\cite{gu2021open} as visual encoders $\phi_i=[\Phi_{clip\_img}(I_i), \Phi_{vild\_img}(I_i)]$,
where image-text-alignment is scored with inner product of image feature and CLIP text feature. %
We also extract the estimated location $p_i=(x_i, y_i, z_i)$ using depth at the center of the image as well as estimated size $r_i$ of the object in $I_i$. Defining the tuple $c_i=(\phi_i,\ p_i,\  r_i)$ as a context element, the collection $\mathcal{C}=\{c_i\}_{i=1...N}$ forms our scene representation.

\subsection{Querying the Representation}\label{sec:alg_query}
To complete a task specified by human instruction, the robot will query the scene representation for relevant information. This is achieved by first parsing natural language instruction into a list of relevant object names, then using the names as keys to query object locations and availability. Finally, we generate executable options based on what's found in the scene, then plan and execute as instructed.

\begin{lstlisting}[language=python,
basicstyle=\ttfamily\footnotesize, 
keywords = {LLM, VLM},  
keywordstyle = {\textbf},
commentstyle=\color{magenta},
label={lst:algobox},
caption={High-level description of \algname + SayCan algorithm. Note we only need to build scene representation once for each scene.},
frame=tb,
escapechar=!]
Input: instruction
if is_new_scene():
    # construct queryable scene representation
    rgbd_images = robot.scene_explore()
    bboxes = roi_proposal(rgbd_images)
    positions, sizes = extract_3d(rgbd_images, bboxes)
    phi = VLM.encode_image(rgbd_images, bboxes)
    nl_map = Context(phi, positions, sizes)
    save_nl_map(nl_map)
else:
    nl_map = load_nl_map()
# extract relevant objects via LLM
objects = LLM.object_proposal(instruction)
# extract text features
queries = VLM.encode_text(objects)
# query the nl_map
object_scores = queries.dot_product(nl_map.Phi)
object_presence, locations = multiview_fusion(object_scores, nl_map)
scene_objects = objects.filter_by(object_presence)
# planning with scene objects information
LLM.plan(instruction, scene_objects)
\end{lstlisting}

\subsubsection{Object proposal}
The core challenge of querying scene information is bridging unstructured natural language input and structured representations. In order to decide what objects to look up in the scene representation, we use few-shot prompting to let LLM actively propose required objects given an instruction. Different from previous work \cite{shah2022lm} that uses LLM to extract names from a sentence, our object proposal is much more demanding in four different ways as we will discuss in Sec.~\ref{sec:exp_proposal}.

In order to achieve a reliable object proposal that addresses four requirements, we introduce example prompts for each case and use the few-shot prompting technique of LLMs to propose them. The few-shot examples can be found on our project website.
\subsubsection{Object Query}
Given a list of object names $\{y_i\}_{i=1...O}$, we then query the scene representation for object locations and availability. This is achieved by finding top k nearest neighbor elements in $\mathcal{C}$ followed by a clustering algorithm to fuse multi-view information. A threshold on a cluster's score determines if the queried object is found.We first define a metric $D: \mathcal{C} \times \mathcal{Y} \rightarrow \mathbb{R}$  where  $\mathcal{Y}$ is the set of possible object names. We use the maximum ensemble of both CLIP and ViLD for the metric D defined below:
\begin{gather*}
D:(\phi_i,\ p_i,\  r_i), \ y_i \mapsto  \max(D_{\text{clip}}, D_{\text{vild}}) \text{, where} \\
D_{\text{clip}} = \langle \Phi_{clip\_img}(I_i) , \Phi_{clip\_text}(I_i)\rangle \\
D_{\text{vild}} = \langle \Phi_{vild\_img}(I_i) , \Phi_{clip\_text}(I_i)\rangle
\end{gather*}
Here we use both CLIP embedding and ViLD embedding because the former detects out-of-distribution objects better while the latter is more robust to common objects as shown in Fig.~\ref{fig:perception}. We can directly take the maximum over the two inner products because both of them are normalized vectors designed to be queried by the inner product CLIP text encoder. Given metric $D$, the top k nearest neighbor elements for object name $y_i$  can be found in the scene representation $\mathcal{C}$. We note that based on the value of $D(c_i, y_i)$, we can impose a threshold to filter out low-confidence detections. These top context elements are associated with ROIs, multiple of which may correspond to the same real-world 3D object instance. We then run a multi-view fusion algorithm to aggregate these context elements into 3d object locations and filter out objects that don't exist according to an aggregated score. Details of the algorithm can be found in Sec. ~\ref{sec:multiview_fusion}.

\begin{figure*}[t]
\setlength{\belowcaptionskip}{-12pt}
  \vspace{0.5em}
    \begin{center}
    \includegraphics[width=0.95\linewidth]{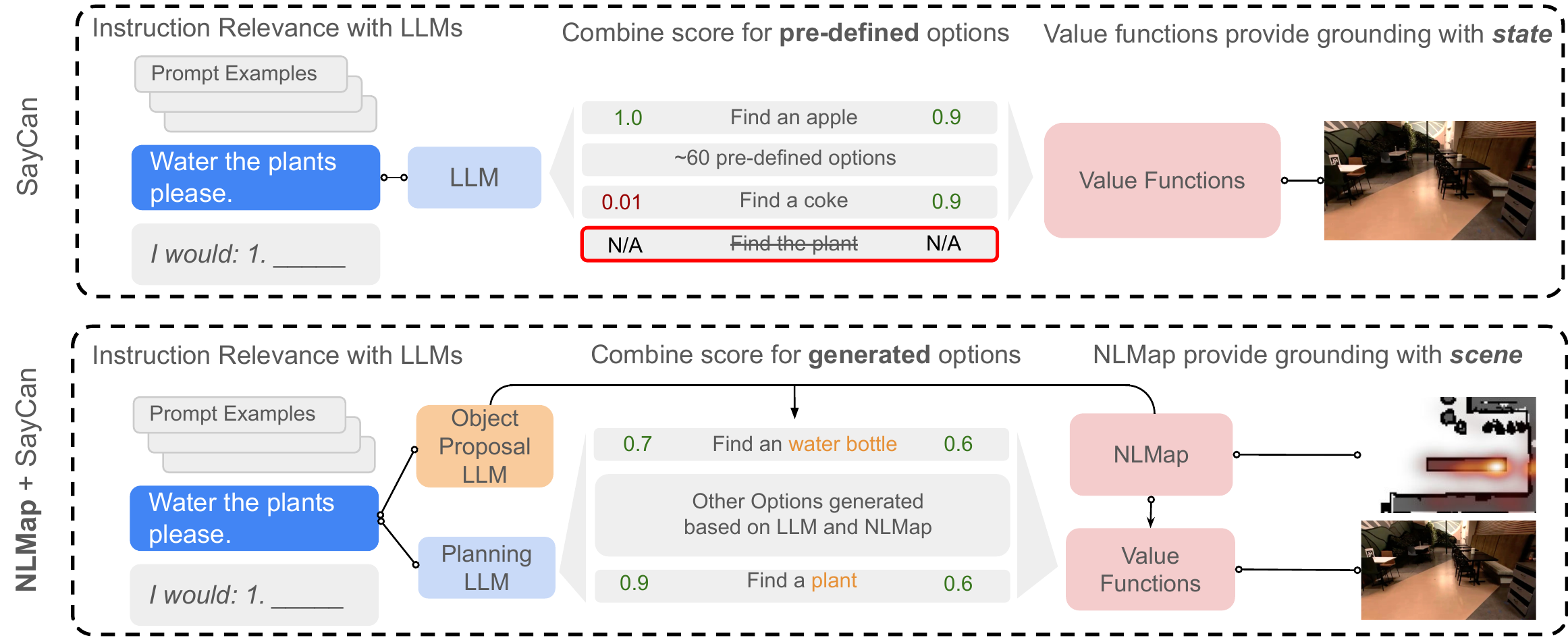}
    \end{center}
    \caption{
        \footnotesize
        \textbf{Comparson of \algname + SayCan with SayCan} \textbf{\emph{SayCan}}: With few-shot prompting, SayCan uses the scoring of a language model to break down a high-level instruction like ``Bring me an apple" to ``1. Find the apple, 2. Pick up the apple, 3. Bring it to you, 4. Put down the apple". Each option from a pre-defined list is scored by an LLM and an affordance prediction module. It natural failed on ``water the plant" task since plant is not contained in the pre-defined options.\textbf{\emph{NLMap + SayCan}} Our method generates a list of relevant objects based on instruction with an LLM module, and then queries the NLMap to filter the object list and get object locations. A list of options is generated based on a template find/pick/place object, and then LLM-based planning module plan over these options.
        \vspace{-2mm}
    }
    \label{fig:saycan_plus}
\end{figure*}

\subsection{Combining \algname and SayCan}\label{sec:alg_saycan}

Our method constructs a scene representation queryable by natural language. Such representation can be connected with LLM-based planners to enable robots to operate in a truly uncontrolled environment. Previously, SayCan\cite{saycan2022arxiv} presents a framework that allows robots to plan and execute in the real world following human instructions. We highlight the difference between our work and SayCan in Fig.~\ref{fig:saycan_plus}. SayCan work as follows: with few-shot prompting, SayCan uses the scoring of a language model to break down a high-level instruction like ``Bring me an apple" to ``1. Find the apple, 2. Pick up the apple, 3. Bring it to you, 4. Put down the apple". Each option from a pre-defined list is scored by an LLM and an affordance prediction module. However, SayCan relies on a hard-coded list of object names, locations, and executable options so its capability is largely limited by the lack of contextual grounding.

\algname makes up this missing component in SayCan. Our object proposal, combined with the object query, generates the relevant object names and locations conditioned on the instruction and the scene. There are two major remaining challenges.
\subsubsection{Generate executable options}
Vanilla SayCan~\cite{saycan2022arxiv} provides a list of skills associated with either 1) navigation policies to hard-coded locations 2) manipulation policies (pick and place) of objects, specified by object names. Given a detected object and its location, we can create a new skill ``find the [object name]'' bound to a navigation policy to that location. This means we can expand a small fixed set of navigation options to infinitely many options. On the other hand, although training manipulation policies for infinitely many objects is beyond the scope of our work, we can still augment the manipulation capability of SayCan by binding all possible references to a manipulable object with the available manipulation policies. This is achieved by finding CLIP nearest neighbor of object names. For example, given discovered objects, we can generate executable options like ``pick up the red can'' and ``pick up a tin of coke''. Our method will bind both of them to the closest manipulation policy ``pick up coke can'' with CLIP. This nearest neighbor query is similar to that used with BERT in~\cite{huang2022language}.

\subsubsection{Ground LLM planner with context} Unlike the setup in SayCan, which assumes all objects in the hard-coded list are present, our method is expected to tackle infeasible instructions, such as instructions involving objects that aren't present. SayCan weakly addresses this problem by grounding plans with local affordance, which is only conditioned on what's directly visible in the field of view rather than what's available in the entire scene. \algname gives us a list of available objects so we can add the missing global contextual grounding to SayCan. This is achieved by modifying the original few shot prompts in SayCan to also condition the plan on discovered objects, expressed in templates like ``Scene: apple, coke can.'' We include both positive examples when necessary objects are all present and negative examples when available objects cannot fulfill the instruction. In the former case, LLM is prompted to plan just like in vanilla SayCan; In the latter case, LLM is prompted to output the terminate signal ``done'' directly, indicating the task is infeasible.

With these components, we can ground SayCan with context awareness. After exploring the scene, when a human gives the robot an instruction, the robot will propose potentially involved objects in the scene and query the gathered scene representation for their locations and availability. \algname then generates executable options, plans with LLM conditioned on what's found and finally executes the plan in the real world under the SayCan framework.

\begin{figure}[h]
\setlength{\abovecaptionskip}{-5pt}
\setlength{\belowcaptionskip}{-12pt}
\begin{center}

    \includegraphics[width=0.99\linewidth]{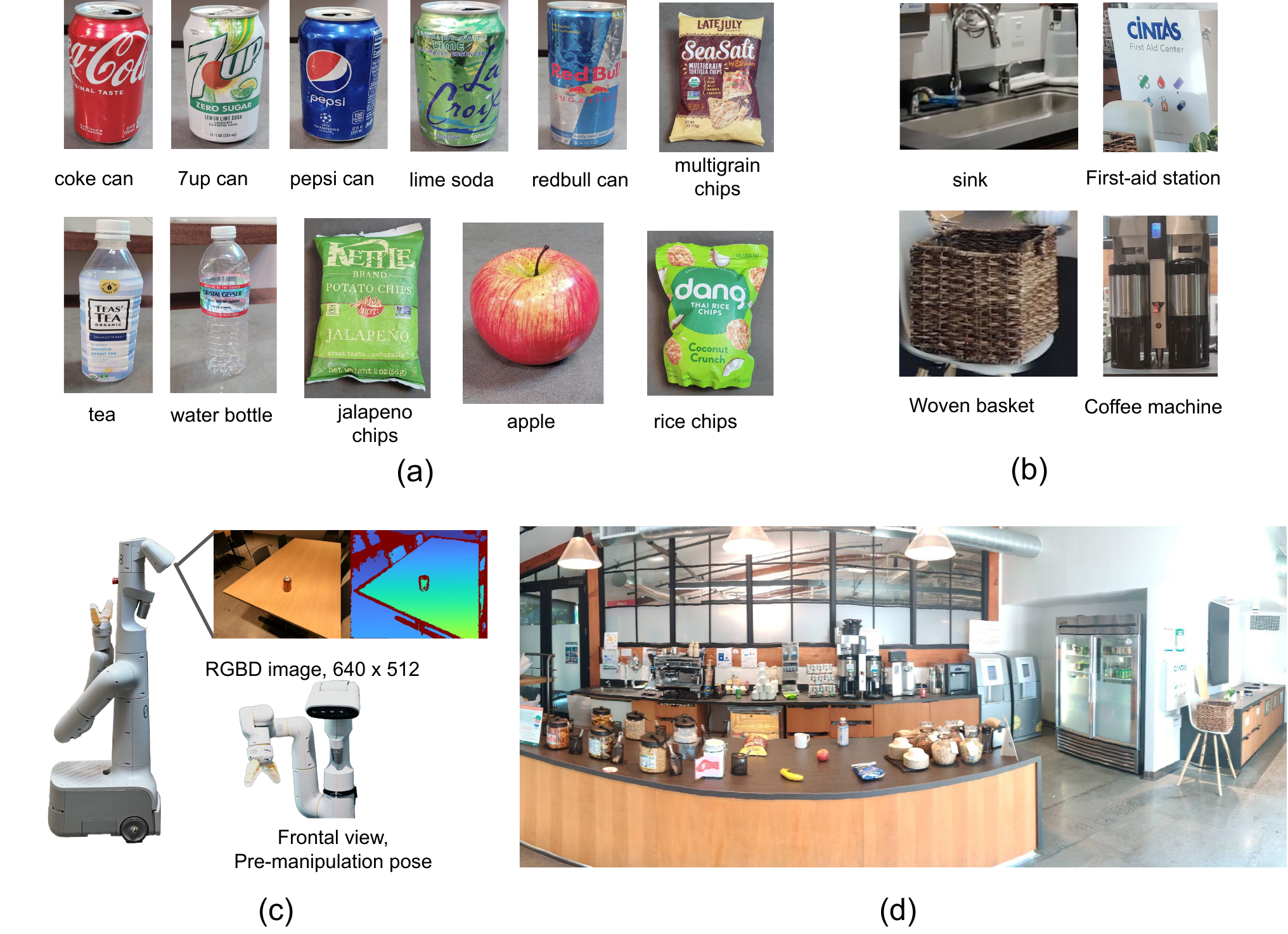}
    \end{center}
    \caption{
        \footnotesize
        (a) a representative subset of objects that are used in manipulation (b) a representative subset of objects used as receptacles (e.g. for the task putting the cup next to the coffee machine) (c) a robot from \href{https://everydayrobots.com/}{Everyday Robots} used in the experiment (d) The scene where we run the experiments, it is a kitchen in an office building.
    \vspace{-2mm}
    }
    \label{fig:exp_setup}
\end{figure}

\section{Experiments}\label{sec:exp}
In this section, we evaluate \algname and its individual components with real-world robotics tasks. We test a robot running \algname in a real office kitchen, as shown in Fig.~\ref{fig:exp_setup}. We test the entire system in an end-to-end setting such that the robot attempts to accomplish tasks specified by humans with natural language. We list a subset of the manipulable objects in Fig.~\ref{fig:exp_setup}(a) receptacle locations in Fig.~\ref{fig:exp_setup}(b). The robot is a mobile manipulator from Everyday Robot, which has a mobile base and a 7-degree-of-freedom arm, as shown in Fig.~\ref{fig:exp_setup}(c). The main sensor is an RGBD camera, which returns $640\times 512$ RGBD images. Similar to SayCan, we use a set of manipulation policies trained from imitation learning and PaLM 540B \cite{chowdhery2022palm} as the LLM for all experiments, due to its good performance on new tasks with few-shot prompting. Throughout this section, all experiments share the same set of hyper-parameters and LLM prompts unless specified otherwise. A full list of test instructions can be found on the project website.

\subsection{Benchmarking \algname + SayCan as a system}\label{sec:exp_saycan}
In this section, we demonstrate our natural language queryable representation can be combined with LLM planners to significantly augment the capability of real robot operation. We choose to combine \algname with SayCan, a recent work that uses LLM planners to let robots plan and execute according to natural language instructions. One of the biggest limitations of SayCan, as stated in Sec.~\ref{sec:problem}, is that it has no global context awareness. By combing our method with SayCan using the method described in Sec.~\ref{sec:alg_saycan}, we free SayCan from a fixed, hard-coded set of objects, locations, or executable options. With \algname, SayCan can now perform a great number of previously unachievable tasks. In addition, we demonstrate that our method allows SayCan to plan with the global context to identify infeasible tasks. We quantitatively evaluate the real robot performance of \algname + SayCan in Table~ \ref{table:e2e} with three sets of benchmarks. We compare our method with a privileged version of SayCan, which uses ground truth perception results in the scene. 

\subsubsection{SayCan tasks}
We hope to understand how much performance will be lost compared to SayCan due to the addition of perception and context-aware planning. Therefore, we benchmark $18$ tasks adopted from $6$ of the $7$ task families from the original SayCan paper with $3$ random tasks from each family (except for Embodiment family). %
Our method achieves a success rate of $66.7\%$ among these tasks compared to the $84\%$ of privileged SayCan. We also tried $2$ tasks with deliberate typos `ppsi'' `chpis''. Our method failed in both instructions with typos, with one failure during object proposal and one failure due to policy binding. With these two typo experiments included, our method achieves an overall success rate of $60\%$ compared to $65\%$ in real robot experiments compared to privileged SayCan that has hard-coded object locations. This shows our \algname maintains a reasonable overall success rate even if multiple components like object proposal, perception, and context-conditioned planning are added. 

\subsubsection{Novel objects}
SayCan relies on a hard-coded list of object names, locations, and executable options. Since the hard-coded set of objects and executable options are finite, SayCan is incapable of performing tasks that involve objects or skills outside these small sets. However, since \algname can propose and detect objects, and generate executable options itself, \algname can be combined with SayCan to execute infinitely many tasks that involve such novel objects as described in Sec.~\ref{sec:alg_saycan}. As shown in Table~\ref{table:e2e}, SayCan fails to plan nor execute any of these tasks while our method achieves a success rate of $80\%$ in the end-to-end execution experiment. It even succeeds in some very out-of-distribution instructions such as ``I want to watch TV, can you get a bottle of tea and put it there'' or ``Show me where is the first aid station''. We note that manipulation policies used in this project are still limited to be with the objects that are visually similar to training objects in~\cite{saycan2022arxiv} and rely on the generalization to slightly out-of-distribution data. Therefore, the novel object names in this experiment are either used for navigation only, or for describing objects that are visually similar to training objects in~\cite{saycan2022arxiv}. Such constraint can be lifted in the future when a general text-conditioned manipulation policy is available but lifting it is beyond the scope of the project.

\subsubsection{Missing Objects}
Vanilla SayCan isn't grounded by what's available in the scene. If a necessary object is removed from the scene, there is no way for SayCan's LLM planner to tell the task is infeasible. With \algname, we can use the method in Sec.~\ref{sec:alg_saycan} to condition SayCan planning on what's actually detected. In this benchmark, we ask \algname + SayCan to perform tasks that require objects not present in the scene. Instructions in the benchmark consist of size $15$ subset of all instructions in the ``novel object'' benchmark since we cannot remove objects like ``first aid station'' from the wall. In a successful run, the robot is expected to not detect an object doesn't exist and output a termination signal immediately in its plan. Our method achieves a success rate of $40\%$ in the missing object setting, where $56\%$ of the total failure cases are due to false positive detections. Although vanilla SayCan will achieve a success rate of zero in comparison, this benchmark still indicates false positive detection is a challenge for context-aware planning.

\begin{table}[]
\setlength{\abovecaptionskip}{-2pt}
\begin{center}
    
\begin{tabular}{lllll}
\toprule
\multirow{2}{*}{Task Family} & \multicolumn{2}{c}{NLMap+SayCan}         & \multicolumn{2}{c}{SayCan$^{*}$}          \\\cmidrule(lr){2-3}\cmidrule(lr){4-5}
                             & \multicolumn{1}{l}{Planning} & Execution & \multicolumn{1}{l}{Planning} & Execution \\ \midrule
SayCan Tasks                 & \multicolumn{1}{l}{ 0.8  }      & {0.6}     & \multicolumn{1}{l}{0.8}      & {\textbf{0.65}}    \\ 
Novel Objects                & \multicolumn{1}{l}{ 0.9  }      & \textbf{0.8}     & \multicolumn{1}{l}{0.0$^{*}$}      & {0.0}     \\ 
Missing Objects              & \multicolumn{1}{l}{  0.67 }      & \textbf{0.4}     & \multicolumn{1}{l}{0.0$^{*}$}      & {0.0}     \\ \bottomrule
\end{tabular}
\end{center}
\caption{\textbf{Planning and execution success rate}. \algname+SayCan shows comparable performance as SayCan on instructions from~\cite{saycan2022arxiv} while enabling new tasks SayCan cannot do before due to its lack of contextual grounding. Planning success rate for \algname + SayCan refers to that of generative planning. ($^{*}$SayCan uses privileged ground truth perception information, thus not able to handle objects out of the pre-defined list.)} 
\label{table:e2e}
\end{table}

\subsection{Benchmarking Object Proposal}\label{sec:exp_proposal}
Object proposal is a foundational component in our framework to parse unstructured instructions into structured object names. We investigate the robustness and generalization capability of object proposal from four perspectives:
\begin{itemize}[wide, labelwidth=!, labelindent=0pt]
    \item Infer objects from implication of the instruction:
    e.g. ``Heat up the taco'' (taco, microwave)
    \item Unstructured crowd-sourced instructions: 
    e.g. ``Redbull is my favorite drink, can I have a one please?'' (redbull, human)
    \item Objects with fine-grained description:
    e.g.  ``turn off  the macbook with yellow stickers'' (macbook with yellow stickers)
    \item Decomposition to proper granularity:
    e.g. ``check out what types of ingredients are available to cook a luxurious breakfast'' (milk,eggs,bacon,bread,butter,cheese,ham,sausage...)
\end{itemize}
A summary of result of each perspective can be found in Table~\ref{table:object_proposal}.

\begin{table}[]
\setlength{\belowcaptionskip}{-12pt}
\setlength{\abovecaptionskip}{-2pt}

\begin{center}
\begin{tabular}{lccc}
\toprule
Task Family              & PaLM540B~\cite{chowdhery2022palm} & PaLM62B & PaLM8B \\ \midrule
Instruction implication  & 0.92     & 0.84    & 0.72   \\ 
Crowd-sourced            & 0.96     & 0.96    & 0.72   \\ 
Detailed description         & 0.72     & 0.8     & 0.6    \\ 
Proper granularity       & 0.6      & 0.2     & 0.133  \\ \bottomrule
\end{tabular}
\end{center}
\caption{Object proposal achieves a very high success rate for all task families except the hardest set ``proper granularity''. The performance on task family ``proper granularity'' sharply declines when we use smaller models while other tasks families witnessed minor decline.}
\label{table:object_proposal}
\end{table}

\subsubsection{Infer objects from implication of the instruction} In previous work \cite{shah2022lm} that use LLM to extract object names from language, all object names are nouns that are directly present in the language input. However, in the real world, humans frequently give instructions that involve objects that have to be inferred from the implication of the task. We test object proposal on $25$ such instructions and evaluate whether proposed objects would complete the task. %
Object proposal achieved a success rate of $92\%$ in $25$ test cases including ``season the steak (salt, pepper)'', ``fillet the fish (fish, knife)''.%
\subsubsection{Unstructured crowd-sourced instructions} Object proposal module is expected to take in instructions from a variety of highly unstructured formats. We evaluate the robustness of our object proposal on a set of $25$ test instructions adopted from crowd-sourced instructions for SayCan. Object proposal achieved a success rate of $96\%$ in this study, including multi-step tasks like ``Move an multigrain chips to the table and an apple to the far counter''. Object proposal succeeded in all $8$ out of $9$  multi-step tasks in this study.
\subsubsection{Reference to objects with fine-grained description} Human instructions often involve reference to objects with fine-grained descriptions. Such descriptions are often important to visually identify a particular instance in the scene. Thus it's important for the object proposal to keep these fine-grained descriptions in its output. We evaluate object proposal on $25$ test instructions that involve fine-grained descriptions by adjectives or clauses. %
The model attains a success rate of $72\%$ in this experiment. The model even succeeded in some complicated descriptions like ``mug in the shape of a donut''. 
\subsubsection{Decomposition to proper granularity} Many instructions require a different level of object proposal granularity. %
Certain tasks can only be accomplished if the object proposal is more fine-grained. We evaluate object proposal on $15$ tasks that require expanding a category mentioned in the instruction. %
Overall, the object proposal achieves a success rate of $60\%$ in this set, indicating that proper granularity is still a hard challenge for LLM due to its multi-modality nature.
\subsection{Benchmarking Object Queries to NLMap}\label{sec:exp_visual}
In this section, we evaluate the open-vocabulary object query module on a list of $50$ common objects in our testing kitchens. %
We run robot exploration and object query in two different kitchen scenes, each with some object deliberately missing. Our method uses both maximum ensemble metric $D$ and multi-view fusion described in Sec.~\ref{sec:alg_query} with $k=4$. We compare this choice with alternative embeddings and metrics like $D_{\text{clip}}$ or $D_{\text{vild}}$. Maximum ensemble metric $D$ without multi-view fusion is also evaluated as a baseline. We have $k=1$ in the above three baselines since no multi-view fusion is happening. As shown in Table~\ref{table:perception}, ViLD and CLIP embedding alone achieves a very low success rate in both environments. As illustrated in Fig.~\ref{fig:perception}, we observe that ViLD embedding detects common objects like cans or apples more reliably while suffering from false negative detection of out-of-distribution objects such as ``first aid station''. On the other hand, CLIP embedding gives us better results on uncommon objects but is less robust for basic objects. Additionally CLIP embeddings better captures features of text and signs. %
Our method uses multi-view fusion in addition to the maximum ensemble. Multiview fusion leads to a slight $2\%$ accuracy increase in scene $1$ but a significant $17\%$ increase in the second scene. This shows that multi-view fusion can help remove outlier observations that produce high likelihood scores but are actually noise by noticing a lack of detection of it from different views. Overall, the perception success rate for our method is $82\%$ and $64\%$ respectively in the two kitchens. Such accuracy is limited by the low resolution and exposure of our robot camera. However, since instructions don't always contain visually ambiguous objects like many in these test queries, perception is still reliable enough as we see in the real robot experiments Sec.~\ref{sec:exp_saycan}.

\begin{table}[]
\setlength{\belowcaptionskip}{-4pt}
\setlength{\abovecaptionskip}{-4pt}
\begin{center}
\begin{tabular}{lcc}
\toprule
Method                            & Scene 1 & Scene 2        \\ \midrule
ViLD embedding                        & 0.6 & 0.47          \\ 
CLIP embedding                        & 0.58 & 0.44          \\
Maximum Ensemble                   & 0.8 & 0.47          \\
Maximum Ensemble + Multiview fusion & \textbf{0.82} & \textbf{0.64}\\ \bottomrule
\end{tabular}
\end{center}
\caption{We ablate different object query methods in two real-world scenes. Both ViLD and CLIP achieve low query success rate but the ensemble of their maximum score as well as our multi-view fusion algorithm provides a significant boost to the query success rate.}
\label{table:perception}
\end{table}

\begin{figure}[h]
\setlength{\belowcaptionskip}{-4pt}
\setlength{\abovecaptionskip}{-4pt}
\begin{center}
    \includegraphics[width=\linewidth]{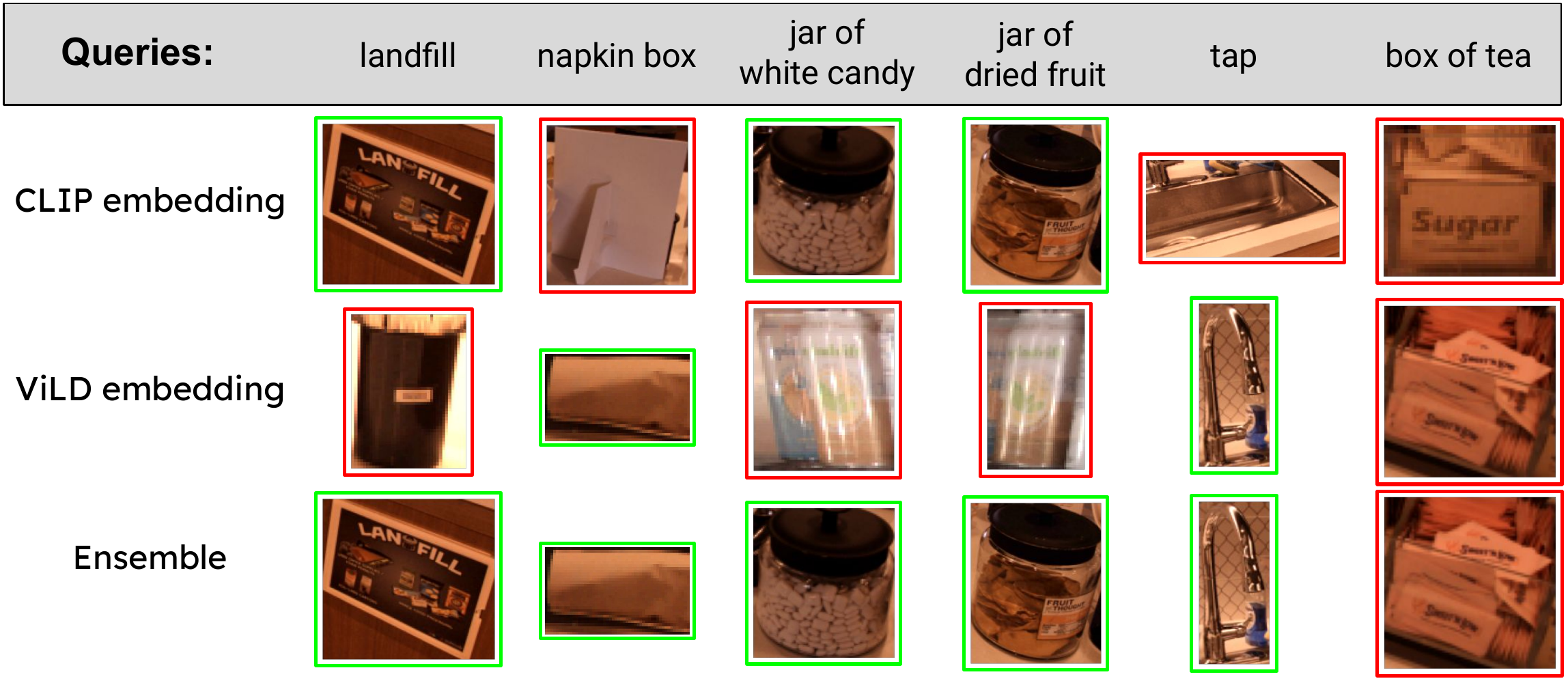}
    \end{center}
    \caption{
        \footnotesize
        \textbf{Comparison of different RoI retrieval method.} We ablate using different features to retrieval RoIs with natural language and found there are unique failure cases with either CLIP or ViLD features, while maximum ensemble of features provide the best results.
    \vspace{-2mm}
    }
    \label{fig:perception}
\end{figure}

\subsection{Benchmarking Context Grounded Planing}\label{sec:exp_generative}
Failures from perception or object proposal are coupled with planning in real robot experiments. In this section, we ablate context-aware LLM planning as a standalone component, assuming correct object proposal and detection. We test LLM planning in a generative way. A generated plan is considered correct if it will accomplish the instruction, is consistent with the available objects, and is executable. We benchmark generative planning with $80$ test cases consisting of $40$ instructions with $2$ set of available objects for each. One set is a positive set that contains all needed objects for the task while the other set is a negative set with some necessary objects missing. To be considered successful, the planner should behave like Vanilla SayCan in the positive set while outputting the terminal signal immediately in the negative set. Our LLM planner, conditioned on available objects using the method described in Sec.~\ref{sec:alg_saycan}, achieves a success rate of $85\%$ and $60\%$ on the $40$ instructions with positive object set and negative set respectively. The performance gap is expected because negation is known to be a hard problem for LLM.

\begin{figure}[h]
\setlength{\belowcaptionskip}{-12pt}
\setlength{\abovecaptionskip}{-4pt}
  \vspace{0.5em}
    \begin{center}
    \includegraphics[width=0.43\textwidth]{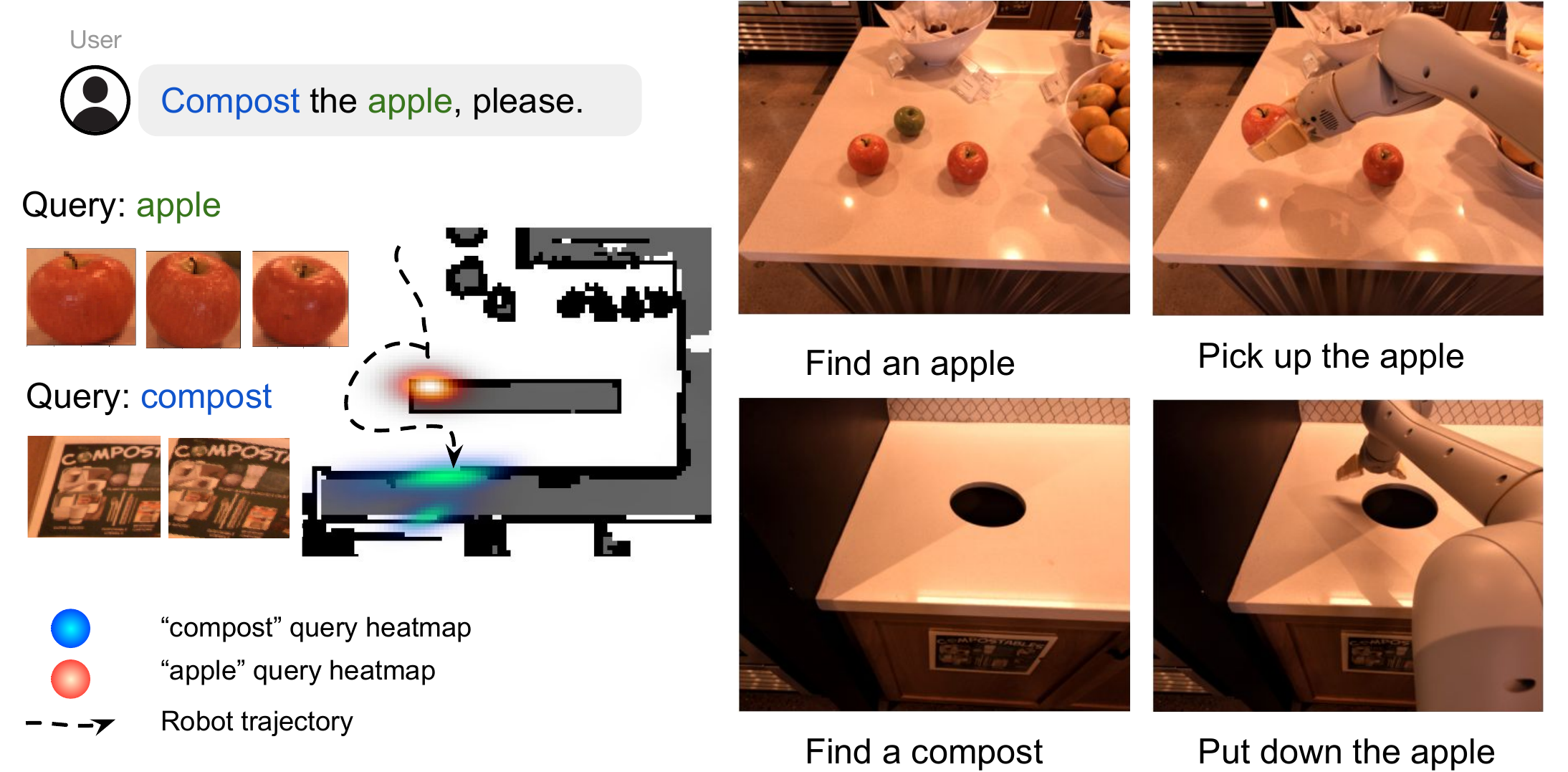}
        \end{center}

    \caption{
        \footnotesize
        Execution trajectory of proposed method on task ``Compost the apple". Note CLIP features allow the robot to understand the sign on the compost bin. The images are from the onboard camera of a robot from \href{https://everydayrobots.com/}{Everyday Robots}.
    }
    \label{fig:taj}
\end{figure}

\section{Conclusions}\label{sec:conclusions}

We integrate \algname, a flexible and queryable spatial semantic representation based on visual-language models including ViLD and CLIP with SayCan. We show that \algname is a flexible scene representation that grounds LLM-based planners in their environments, significantly improving long-horizon planning via natural language instructions in open-worlded domain, enabling new tasks prior state-of-the-art algorithms failed to address. 

\textbf{Future work.} Currently, \algname only handles a static scene representation without dynamic objects and human, which we will leave this for future work. All the modules used in \algname + SayCan is pre-trained and deployed zero-shot. It is a great advantage but we  hope to fine-tune them for better performance. Additionally, we will look into efficient exploration algorithms to speed up the creation of scene representation.

\section*{Acknowledgements}
Special thanks to Arjun Majumdar, Andy Zeng and Karol Hausman for helpful discussions; we thank Peng Xu and Xiran Liu for helpful feedbacks on writing.

\bibliography{cite}
\bibliographystyle{IEEEtran-short}

\newpage

\section*{APPENDIX}

\subsection{Context-aware SayCan Algorithm}
Our context-aware SayCan algorithm is similar to  \cite{saycan2022arxiv}, it expands the last line \texttt{LLM.plan(instruction, scene\_objects)} in Listing~\ref{lst:algobox}. Compared to the original SayCan \cite{saycan2022arxiv}, our context-aware version needs a list of detected object names $\mathcal{M}$, along with a list of template functions $\mathcal{F}$ as extra input. A template function maps an object name to an option name such as $x\longrightarrow ``\text{pick up } [x]''$. We note that the template function is used here because training manipulation policies beyond pick-and-place are beyond the scope of our project. If we have a language-conditioned policy in the future, we don't need to use template functions anymore. Trusting LLM for new options will suffice in that case. A full pseudo-code can be found in Algo ~\ref{algo:contextsaycan}.

\begin{algorithm}
\caption{Context-Aware SayCan}\label{alg:main}
\label{algo:contextsaycan}
\begin{algorithmic}[1]
\State \textbf{Input}: A high level instruction $i$, a list of detected scene object names $\mathcal{M}$, a list of template functions $\mathcal{F}$, state $s_0$, and a set of skills $\Pi$ and their affordance functions $V_\Pi$ along with their language descriptions $d_\Pi$. 

\State $l_{\mathcal{A}} \gets$ [``done'']
\State translate$ \gets \{\}$ 
\For{o $\in \mathcal{M}$}
    \For{$f \in \mathcal{F}$}
        \State $\ell_{\mathcal{A}}$.append($f(o)$)\Comment{Create executable options}
        \State $\pi_{nn} \gets \text{argmax}_{\pi \in \Pi}\langle \text{clip}(d_\pi), \text{clip}(f(o))\rangle$
        \State translate[$f(o)$] =  $\pi_{nn}$
        \Comment{Bind options to policies}
    \EndFor
\EndFor

\State $n \gets 1$
\While{$\ell_{\mathcal{A}_{n-1}} \neq \text{``done''}$}
\State $\mathcal{Q} = \emptyset$
\For{$a \in \mathcal{A}$ and $\ell_{a} \in \ell_{\mathcal{A}}$}
\State $\pi = \text{translate}[\ell_a]$
\State $q_{a}^{\mathrm{LLM}} = p(\ell_{a} | i, \mathcal{M}, \ell_{a_{n-1}}, ..., \ell_{a_1})$ \Comment{LLM score}
\State $q_{a}^{\mathrm{affordance}} = V_{\pi}(s_n)$ \Comment{Affordance}
\State $q_{a}^{\mathrm{combined}} = q_{a}^{\mathrm{affordance}} q_{a}^{\mathrm{LLM}}$
\State $\mathcal{Q} = \mathcal{Q} \cup q_{a}^{\mathrm{combined}}$
\EndFor
\State $a_n = \text{argmax}_{a \in \mathcal{A}}{\mathcal{Q}}$
\State $\pi_n=\text{translate}[\ell_{a_n}]$
\State Execute $\pi_n(s_n)$ in the environment, updating state $s_{n+1}$
\State $n = n+1$
\EndWhile
\end{algorithmic}
\end{algorithm}

\subsection{Multi-view fusion algorithm}
\label{sec:multiview_fusion}
In this section, we describe details of the multi-view fusion algorithm mentioned in Sec.~\ref{sec:alg_query}. In the gathered scene representation $\mathcal{C}$, multiple context elements may be associated with the same object. Each context element $c_i$ contains an estimation of object centroid $p_i$ and along with a object width $r_i$. To simplify formulation, we use cylindrical bounding volumes to model 3d objects. We create such bounding boxes with center $p_i$ and radius $r_i$ in an upright position. Given each queried object name $y_i$, we can quickly narrow down bounding box candidates by finding the top k nearest neighbors with metric $D$. We now have a problem similar to post-processing in object detection - for each real object instance, we may have overlapping bounding box predictions, which are supposed to be aggregated together. In computer vision, this is achieved by the NMS algorithm that group predictions based on the intersection over union(IOU) of the bounding box followed by keeping only the bounding box with the highest confidence in each group. We made three major changes to the NMS algorithm by noticing the special structure of our problem. 

First, since our bounding volumes are not cubes, IOU is hard to compute. We instead use KL divergence of Gaussian distributions to model. For each cylindrical bounding box $(p_i, r_i)$ with a circular projection on the 2d plane, define Gaussian distribution $G_i=\mathcal{N}(p_{i}, \alpha \cdot r_{i})$. The 2d Gaussian will have its center at the estimated centroid and standard deviation proportional to the width of the object. KL divergence measures how different two distributions are so it acts like the IOU for gaussian distributions. When estimations have very different centers or sizes, they will be considered to correspond to two different object instances by our algorithm. Second, different from the setup in 2d object detection, different estimations of the same object in our problem are considered valid, independent data points that contribute to a better estimation of object location. Therefore, we don't discard non-maximum estimations in each clustered group, but rather use their score as importance weights to derive the final estimation through weighted average. Third, bounding boxes are directly filtered out based on a threshold on confidence score in 2d detection. In our setup, we give confidence scores a bonus based on how many elements there are by noticing available objects should be detected from multiple view points. 

We then offer a formal algorithm box for multi-view fusion in Algo \ref{algo:multiviewfusion}. Given object name $y$, we can use metric $D$ to score each context element in $\mathcal{C}$ and find the top k ones. Denote the indices of top k context elements as $\mathcal{K}$, sorted in descending order by score. For each context element $c_i=(\phi_i,\ p_i,\  r_i)$ , define Gaussian distribution $\mathcal{N}_i=\mathcal{N}(p_{i}, \alpha \cdot r_{i})$. In our experiments, we choose the monotonic increasing function $f$ to be in the form $f(x)=1+t -\frac{t}{x}$ where $t$ is some hyper-parameter.

\begin{algorithm}

\begin{algorithmic}[1]
\caption{Multi-view Fusion in \algname}\label{alg:cluster}
\label{algo:multiviewfusion}
\State \textbf{Input}: Sorted indices $\mathcal{K}$, Scores $S$ for context elements, Gaussian distributions for context elements $N$, KL threshold $\lambda$, score threshold $\beta$, monotonic increasing function $f$. \newline

\State Groups $\gets [[\mathcal{K}[0]]]$
\For {i $\in \mathcal{K}$}
    \For {j $\in \mathcal{K}$}
        \If{$\forall$ G $\in$ Groups, $\forall z \in$ G , $i\ne z \wedge j\ne z$}
            \If{$KL(\mathcal{N}_i$, $\mathcal{N}_j)>\lambda$}
                \State Groups$[-1]$.append$(j)$
            \Else
                \State Groups.append$([i])$
            \EndIf
        \EndIf
    \EndFor
\EndFor
\State P $\gets []$
    \For{G $\in $ Groups}
        \If{$G[0]\cdot f(|G|)>\beta$}
            \State P.append$(\frac{\sum_{i\in G} p_i \exp(S_i)}{\sum_{i\in G} \exp(S_i)})$
        \EndIf
    \EndFor
\State \Return{P}
\end{algorithmic}
\end{algorithm}

The algorithm then outputs clustered locations for objects queried by name $y$.

\subsection{Prompt used for object proposal and for planning}
\input{prompt}
\subsection{Object proposal experiment task list}
\input{obj_proposal_task}
\subsection{Robot experiment task list}
\input{task_list}

\subsection{Additional qualitative experiment results}
\begin{figure*}[h]
\centering
\begin{subfigure}[b]{1.0\textwidth}
         \centering
        \includegraphics[width=0.85\textwidth]{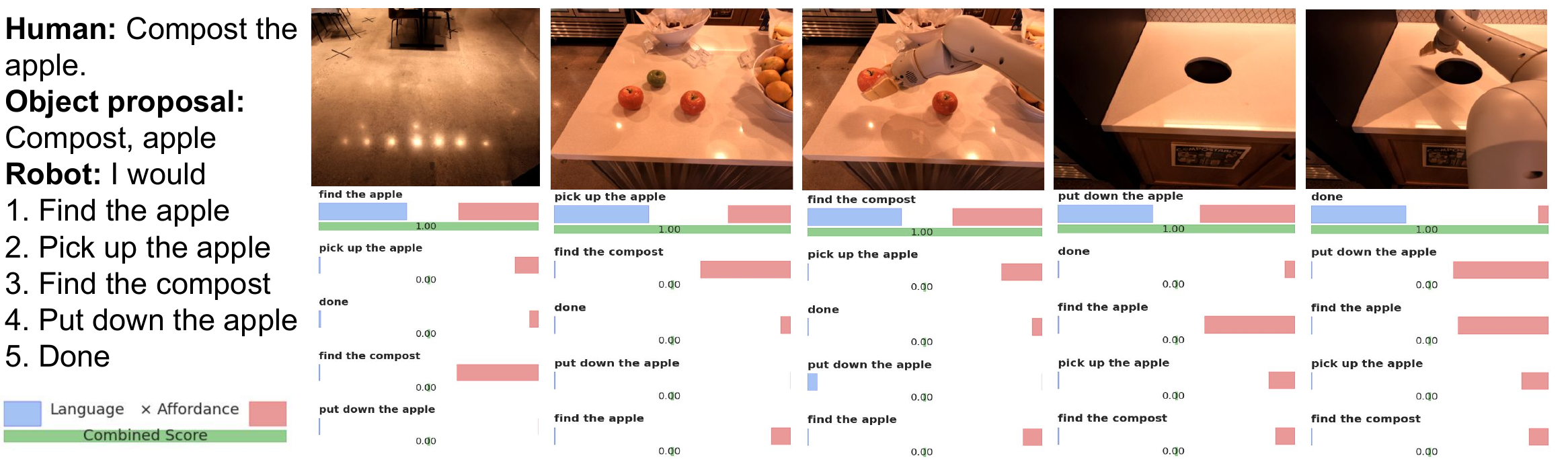}
         \caption{}
\end{subfigure}
\begin{subfigure}[b]{1.0\textwidth}
         \centering
        \includegraphics[width=0.85\textwidth]{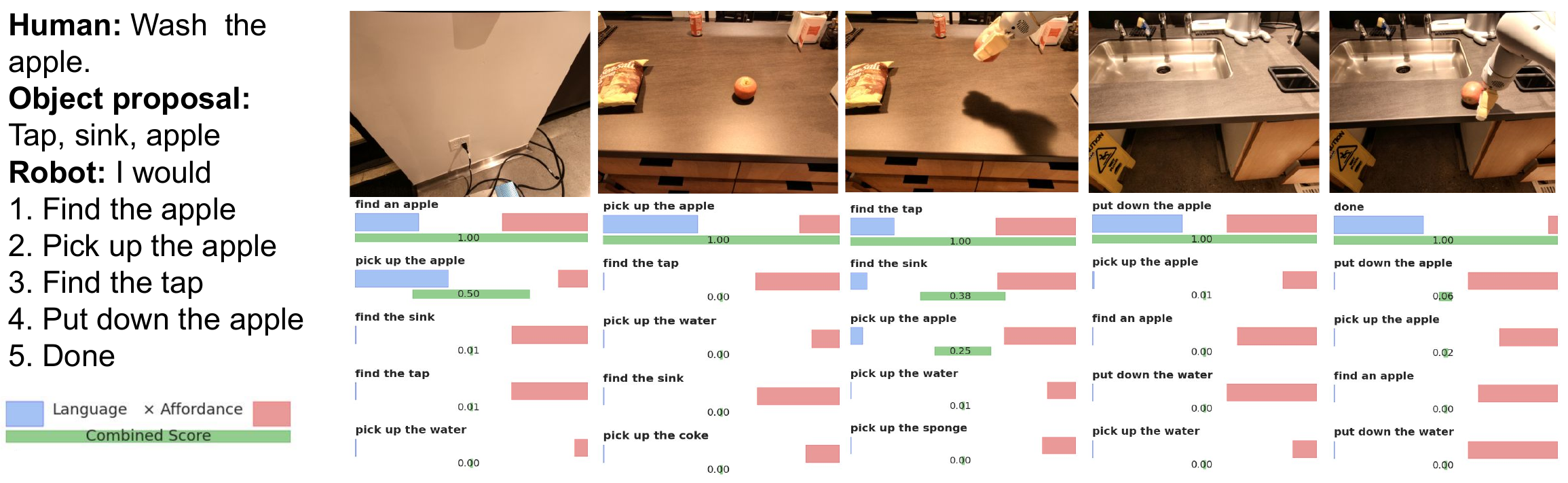}
         \caption{}
\end{subfigure}
\caption{\textbf{Examples execution of of \algname + SayCan on two tasks.} (a) The robot is able to find apple and compost, based on CLIP and ViLD features, and plan a sequence to put the apple in the compost. (b) Based on the instruction ``Wash the apple", the robot is able to bring the apple to the tap. Turning on the tap is out of scope of this project and left for future work. The images are from the onboard camera of a robot from \href{https://everydayrobots.com/}{Everyday Robots}.\label{fig:add1}}
\end{figure*}

\begin{figure*}[h]
\centering
\begin{subfigure}[b]{1.0\textwidth}
         \centering
        \includegraphics[width=0.65\textwidth]{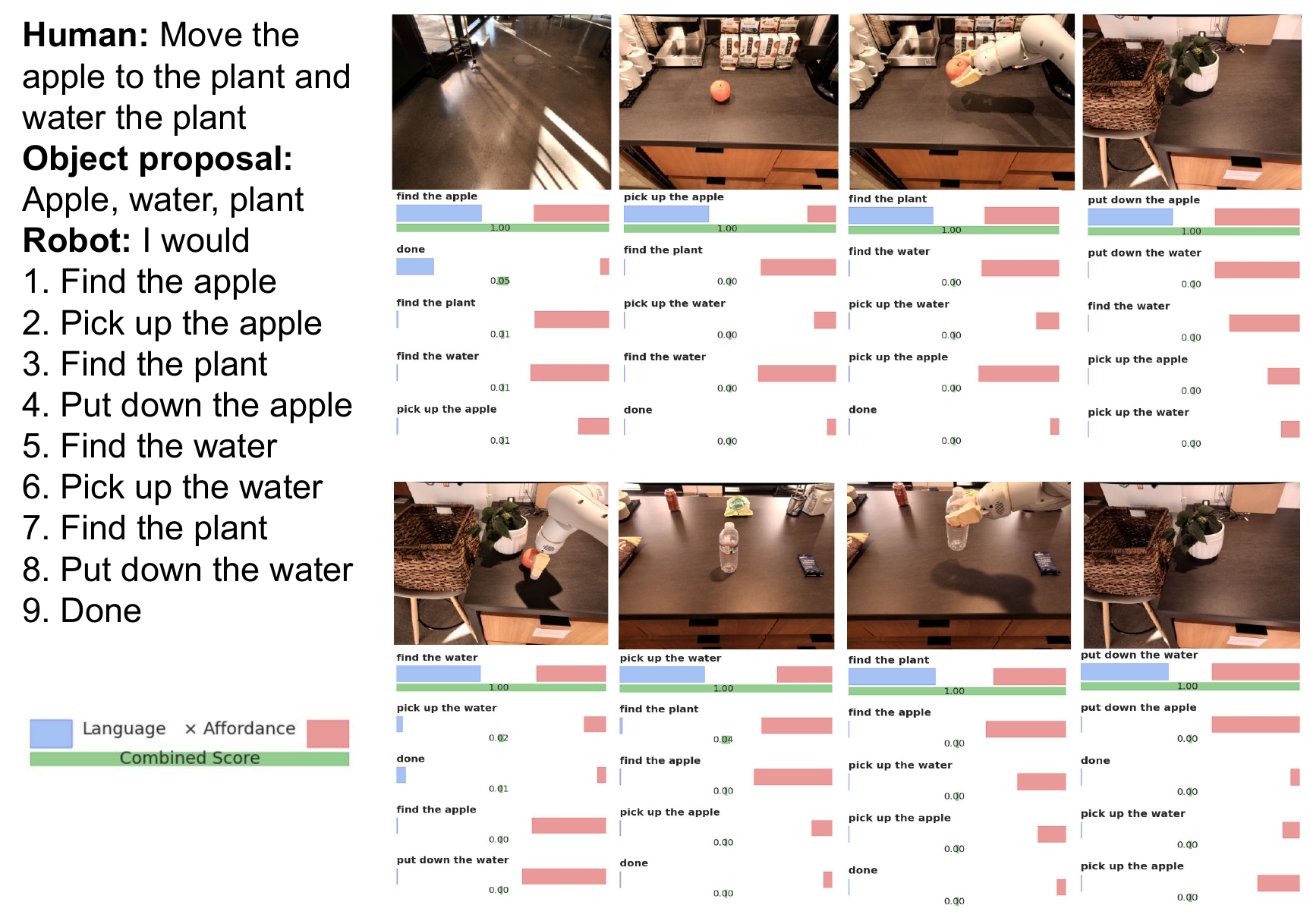}
         \caption{}
\end{subfigure}
\begin{subfigure}[b]{1.0\textwidth}
         \centering
        \includegraphics[width=0.65\textwidth]{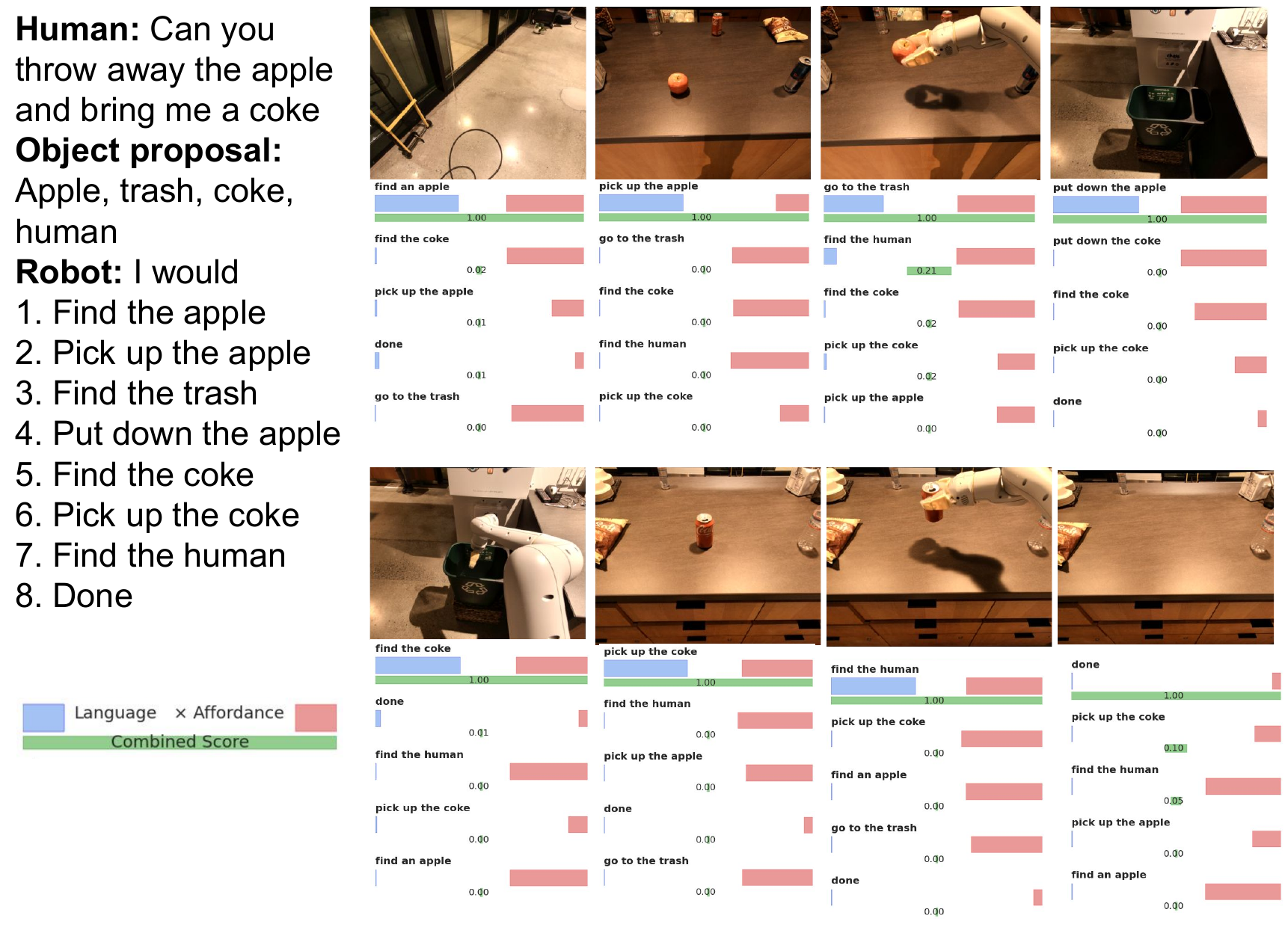}
         \caption{}
\end{subfigure}
\caption{\textbf{Examples execution of \algname + SayCan on long horizon tasks.} The proposed method is able to propose objects and plan a sequence for long horizon tasks ((a) has 9 steps and (b) has 8 steps).  The images are from the onboard camera of a robot from \href{https://everydayrobots.com/}{Everyday Robots}.\label{fig:add2}}
\end{figure*}

\begin{figure*}[h]
\centering
\begin{subfigure}[b]{0.24\textwidth}
         \centering
        \includegraphics[width=\textwidth, trim={2cm 2cm 1cm 1cm},clip]{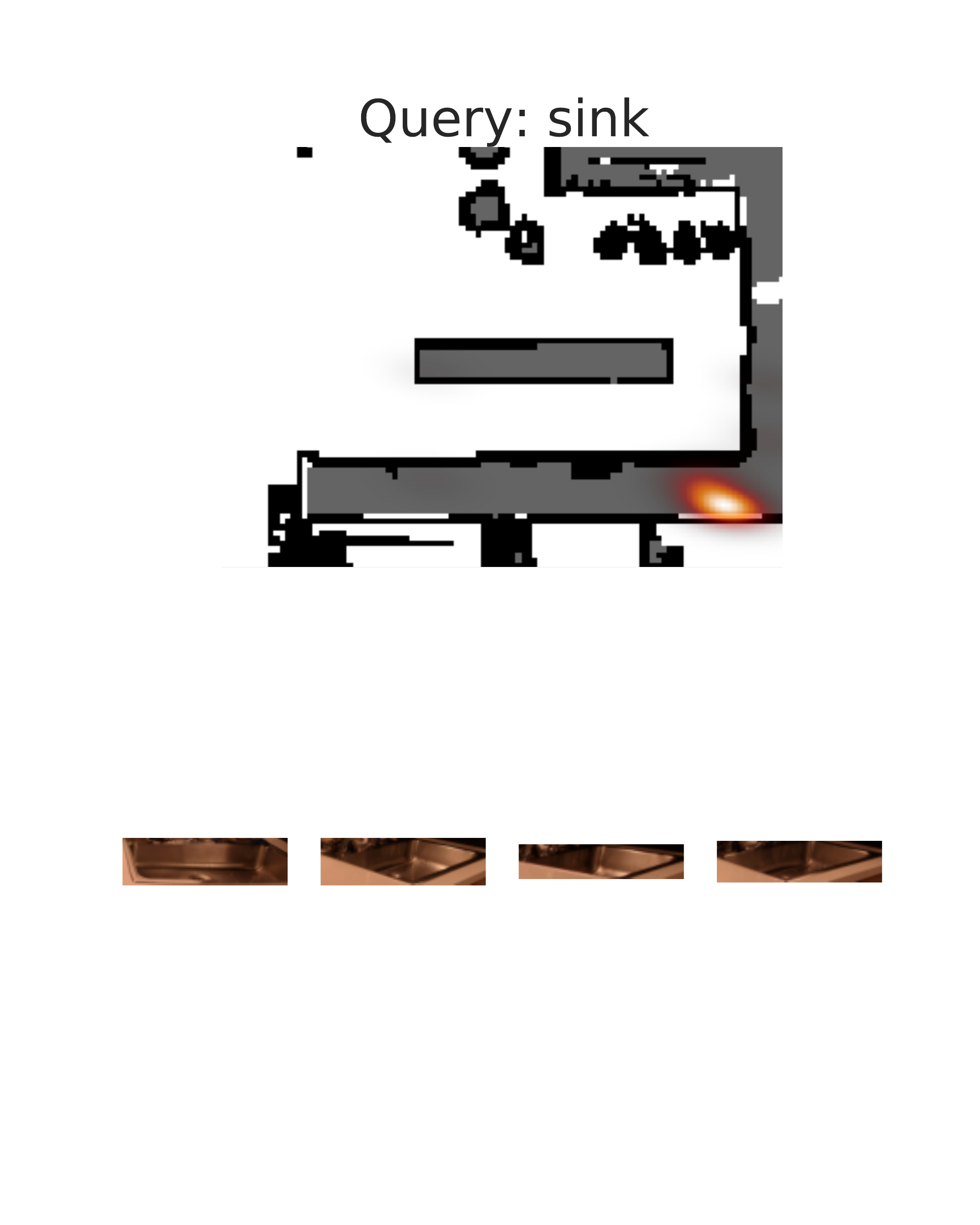}
         \caption{}
\end{subfigure}
\begin{subfigure}[b]{0.24\textwidth}
         \centering
        \includegraphics[width=\textwidth, trim={2cm 2cm 1cm 1cm},clip]{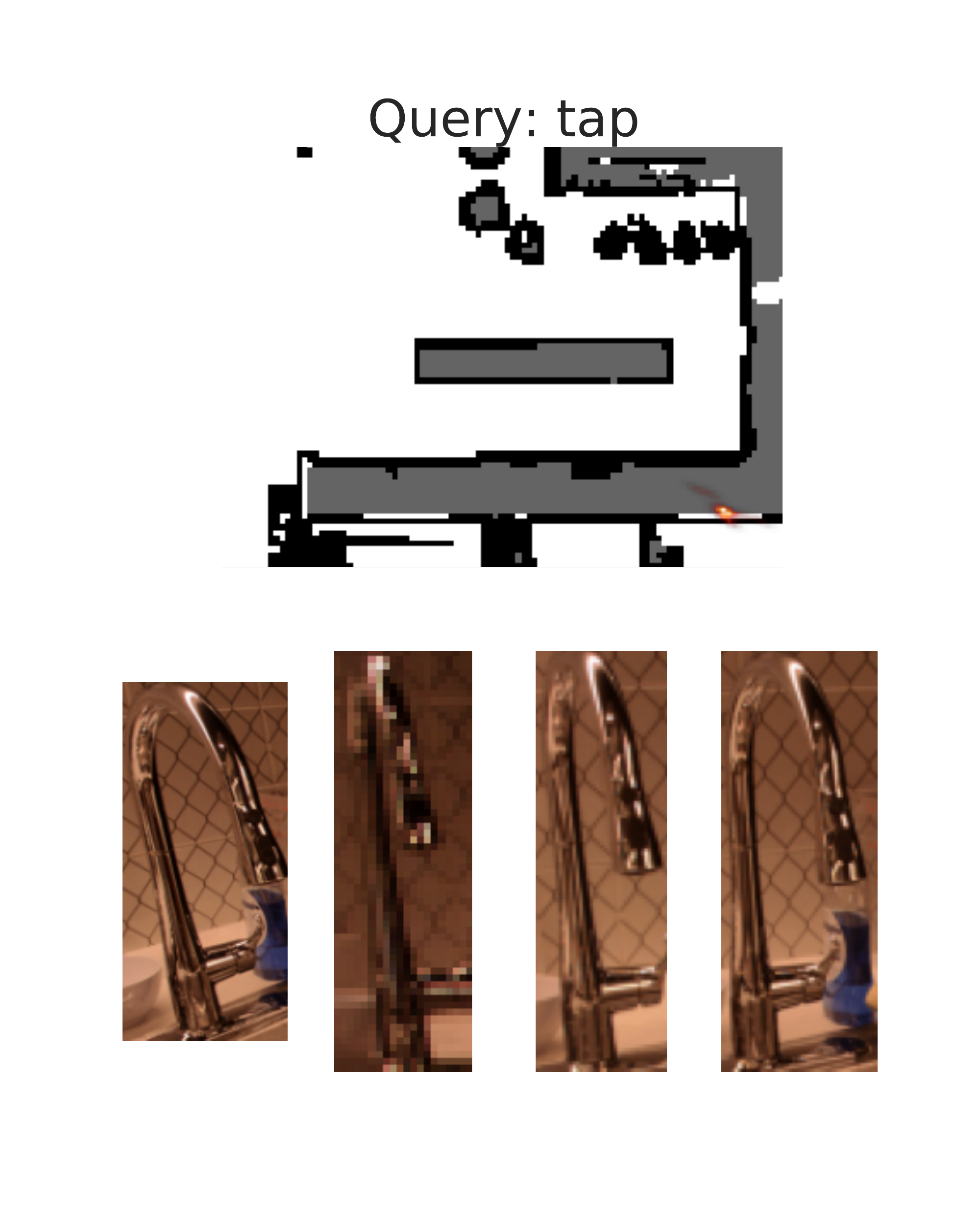}
         \caption{}
\end{subfigure}
\begin{subfigure}[b]{0.24\textwidth}
         \centering
        \includegraphics[width=\textwidth, trim={2cm 2cm 1cm 1cm},clip]{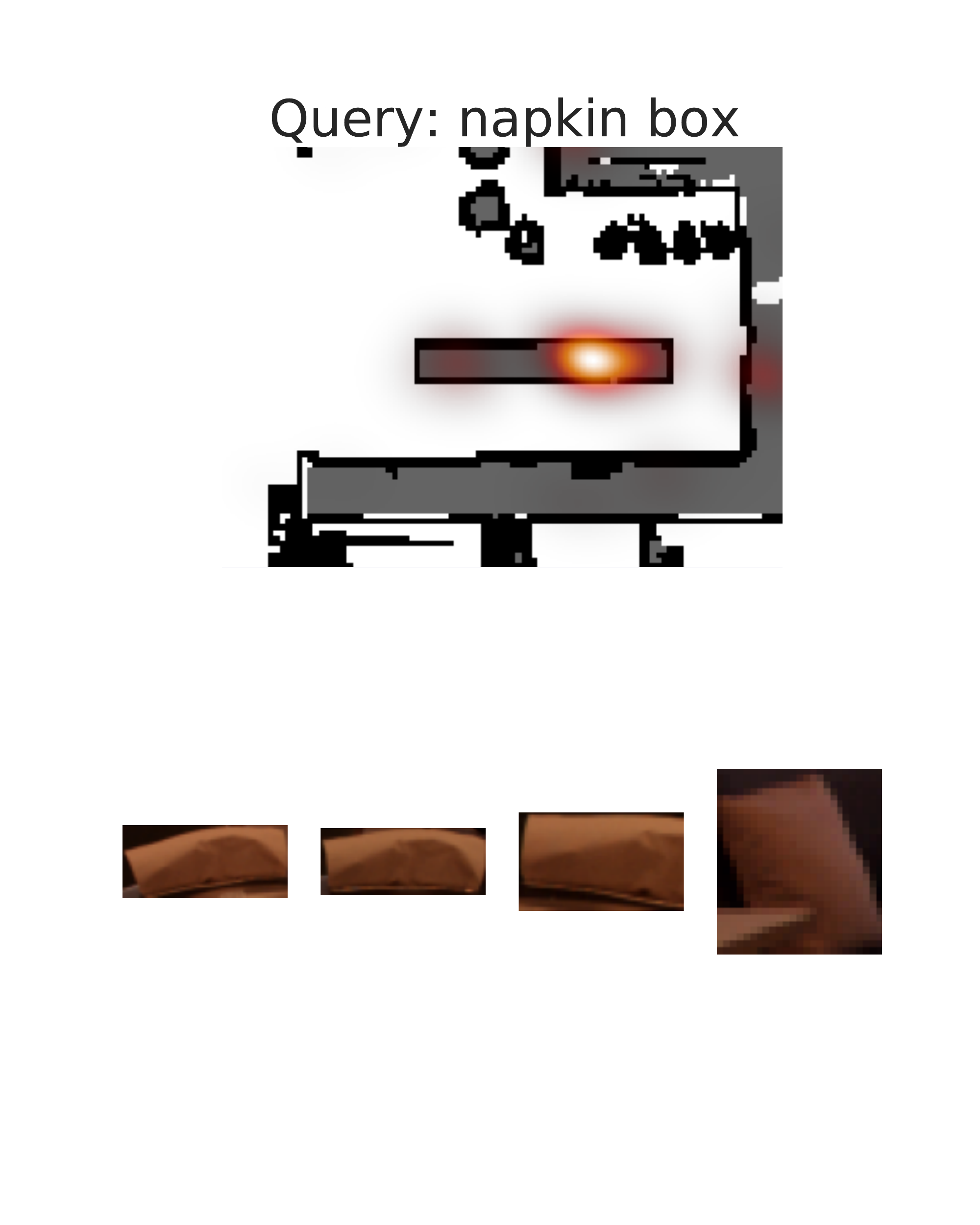}
         \caption{}
\end{subfigure}
\begin{subfigure}[b]{0.24\textwidth}
         \centering
        \includegraphics[width=\textwidth, trim={2cm 2cm 1cm 1cm},clip]{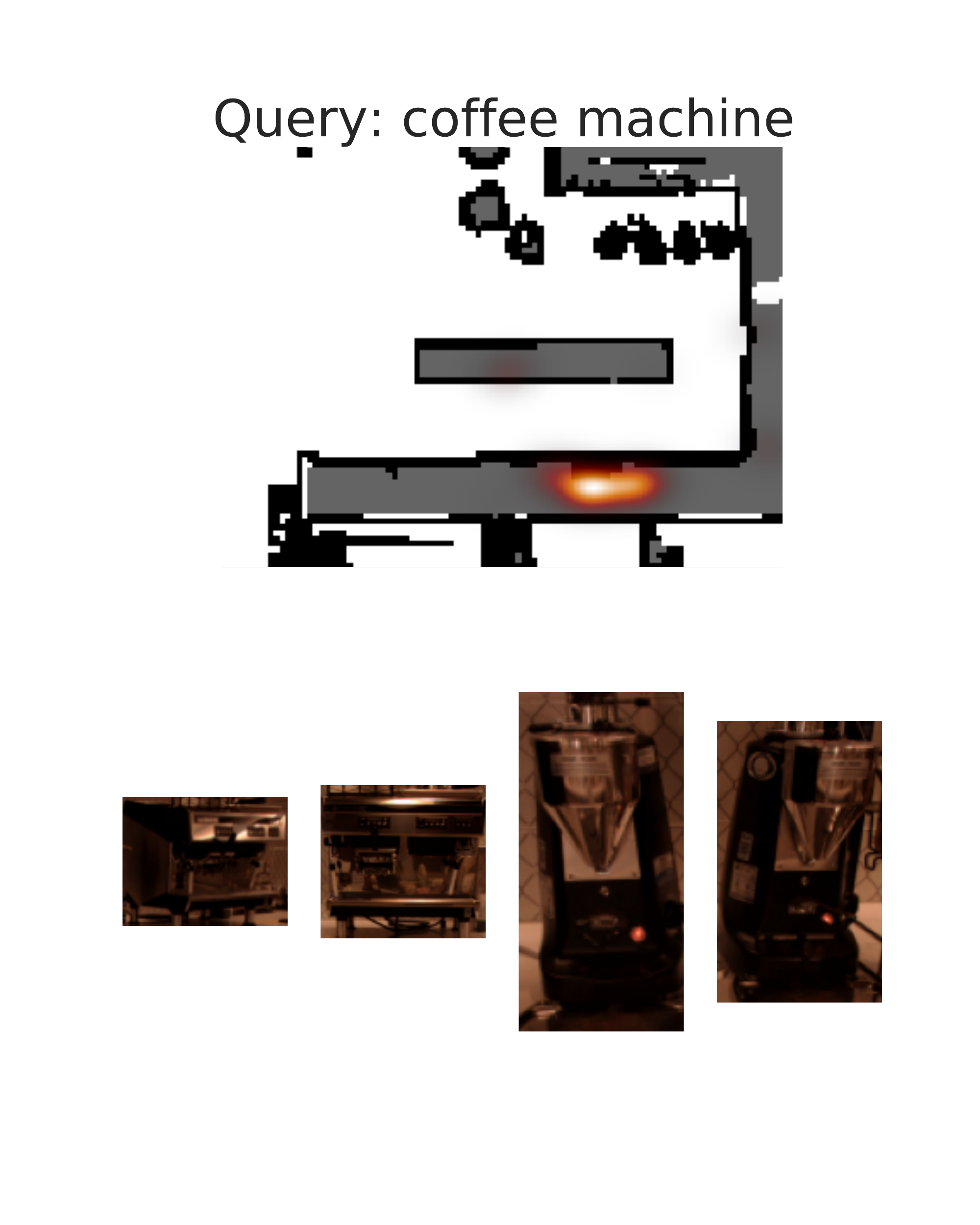}
         \caption{}
\end{subfigure}

\begin{subfigure}[b]{0.24\textwidth}
         \centering
        \includegraphics[width=\textwidth, trim={2cm 2cm 1cm 1cm},clip]{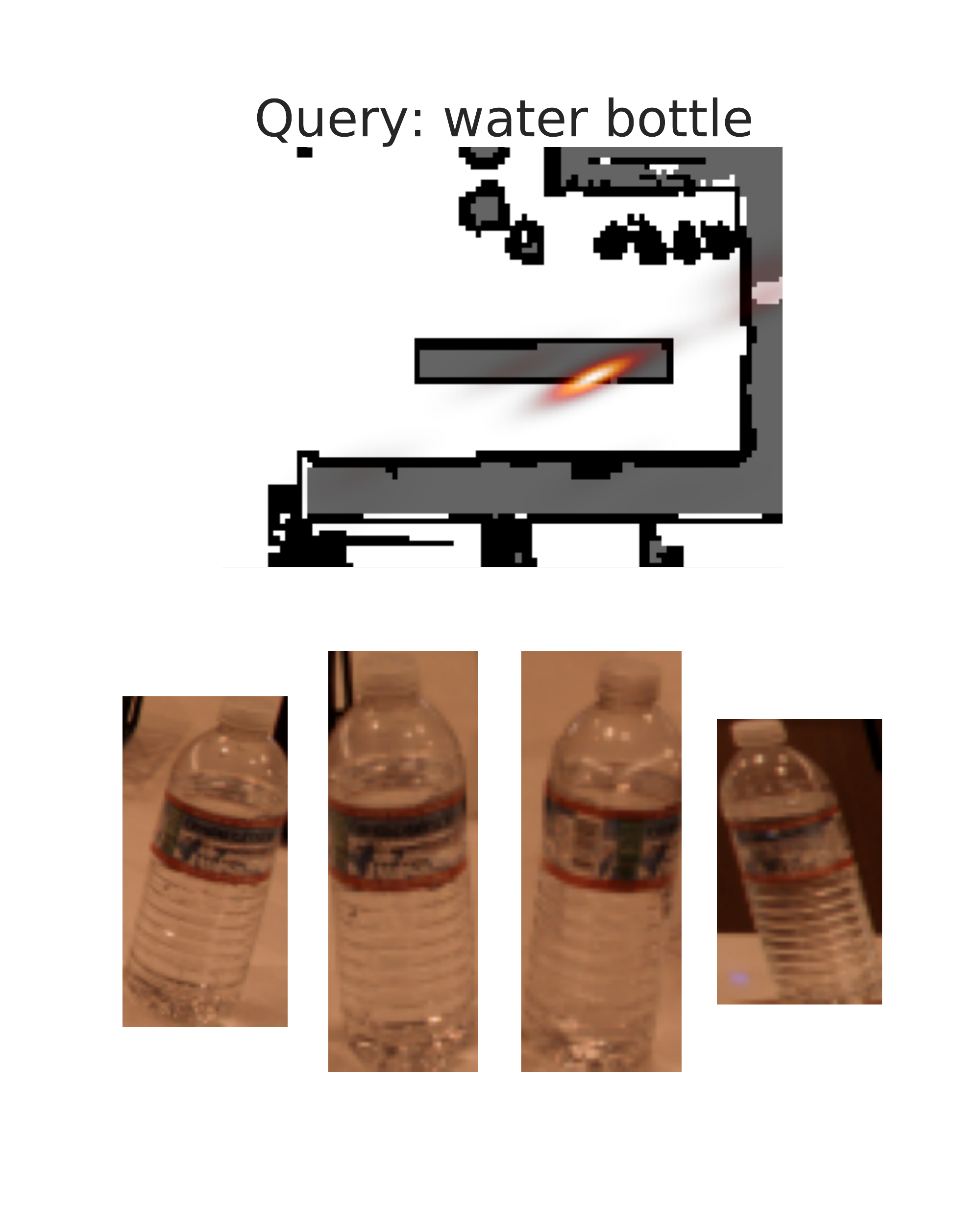}
         \caption{}
\end{subfigure}
\begin{subfigure}[b]{0.24\textwidth}
         \centering
        \includegraphics[width=\textwidth, trim={2cm 2cm 1cm 1cm},clip]{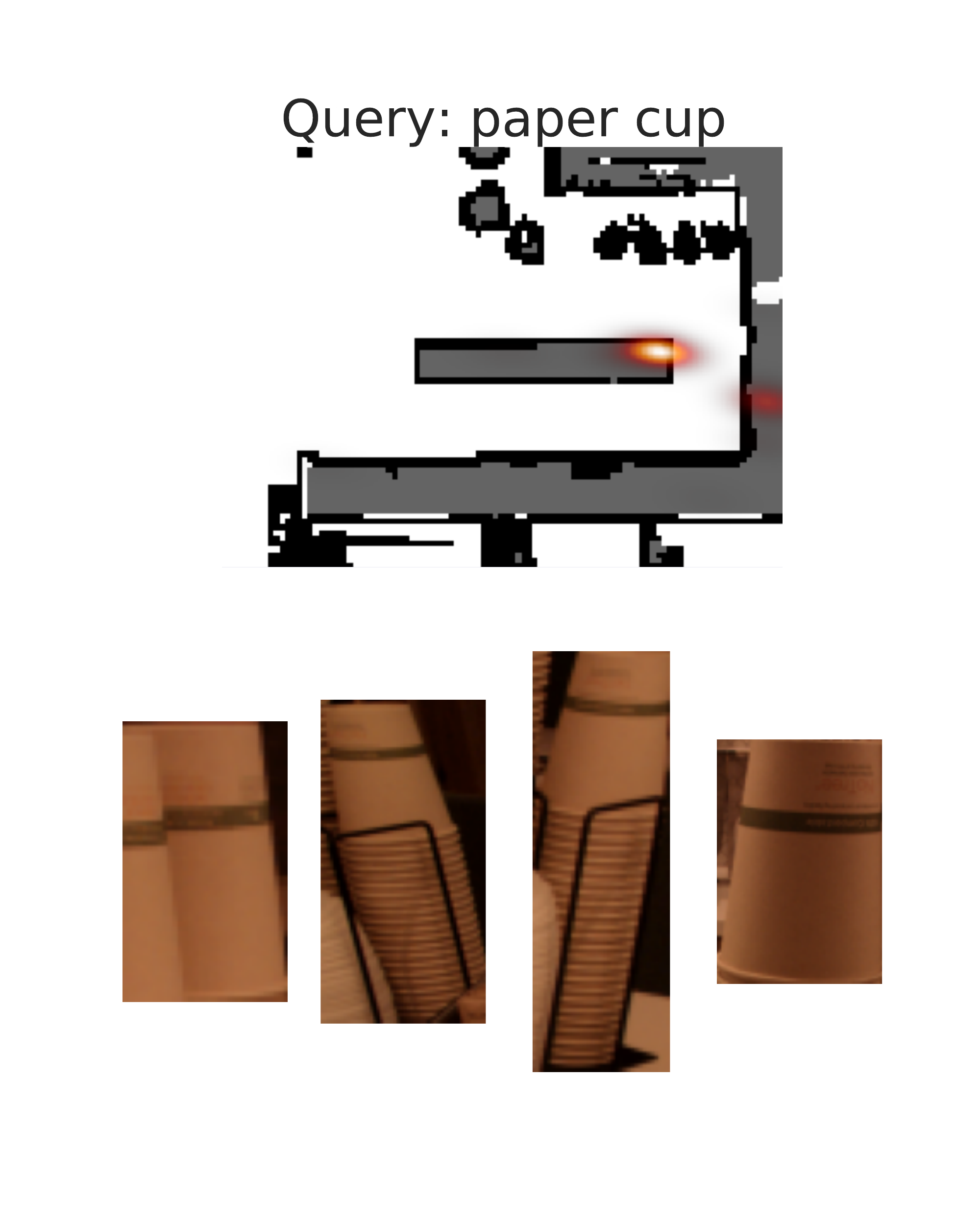}
         \caption{}
\end{subfigure}
\begin{subfigure}[b]{0.24\textwidth}
         \centering
        \includegraphics[width=\textwidth, trim={2cm 2cm 1cm 1cm},clip]{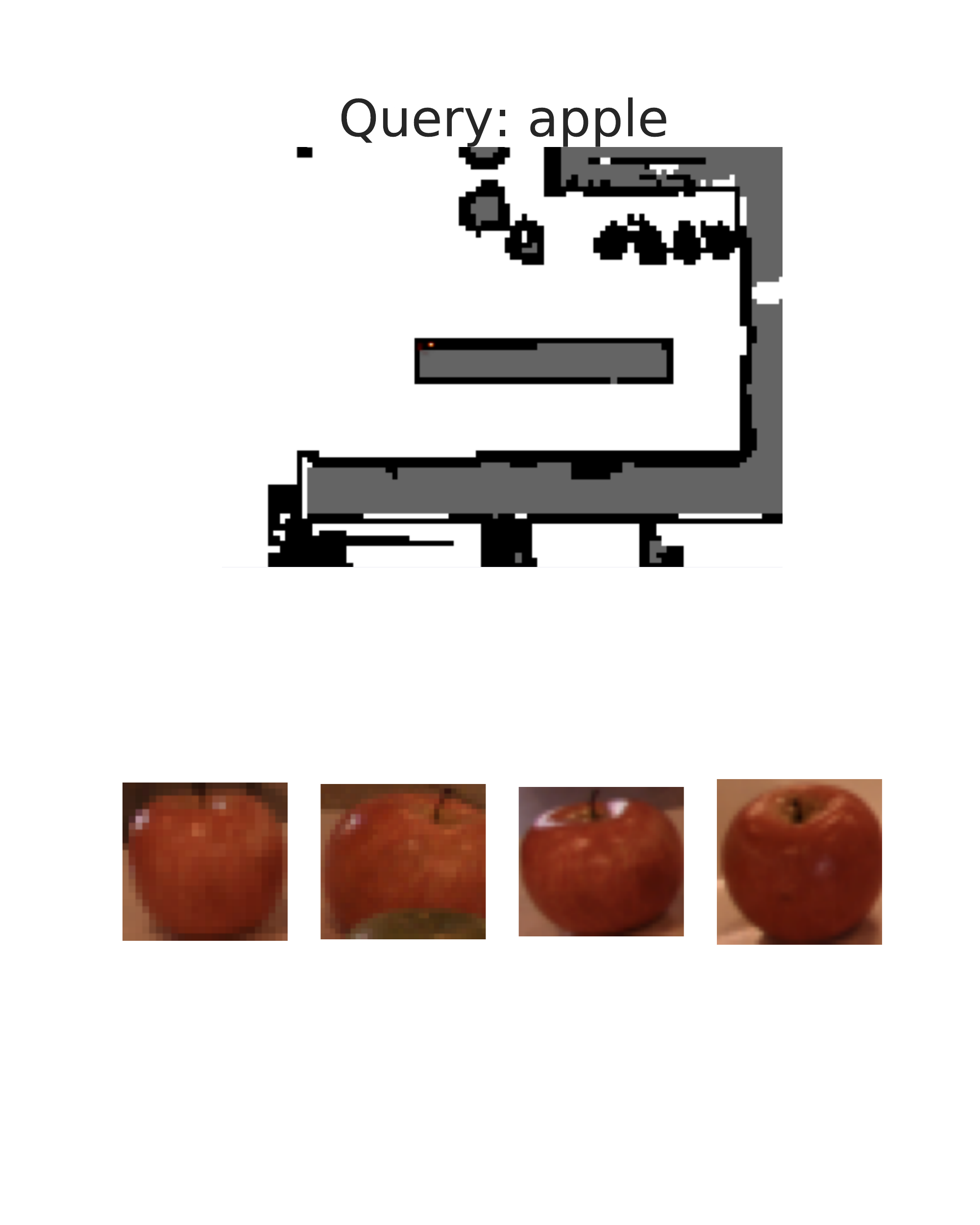}
         \caption{}
\end{subfigure}
\begin{subfigure}[b]{0.24\textwidth}
         \centering
        \includegraphics[width=\textwidth, trim={2cm 2cm 1cm 1cm},clip]{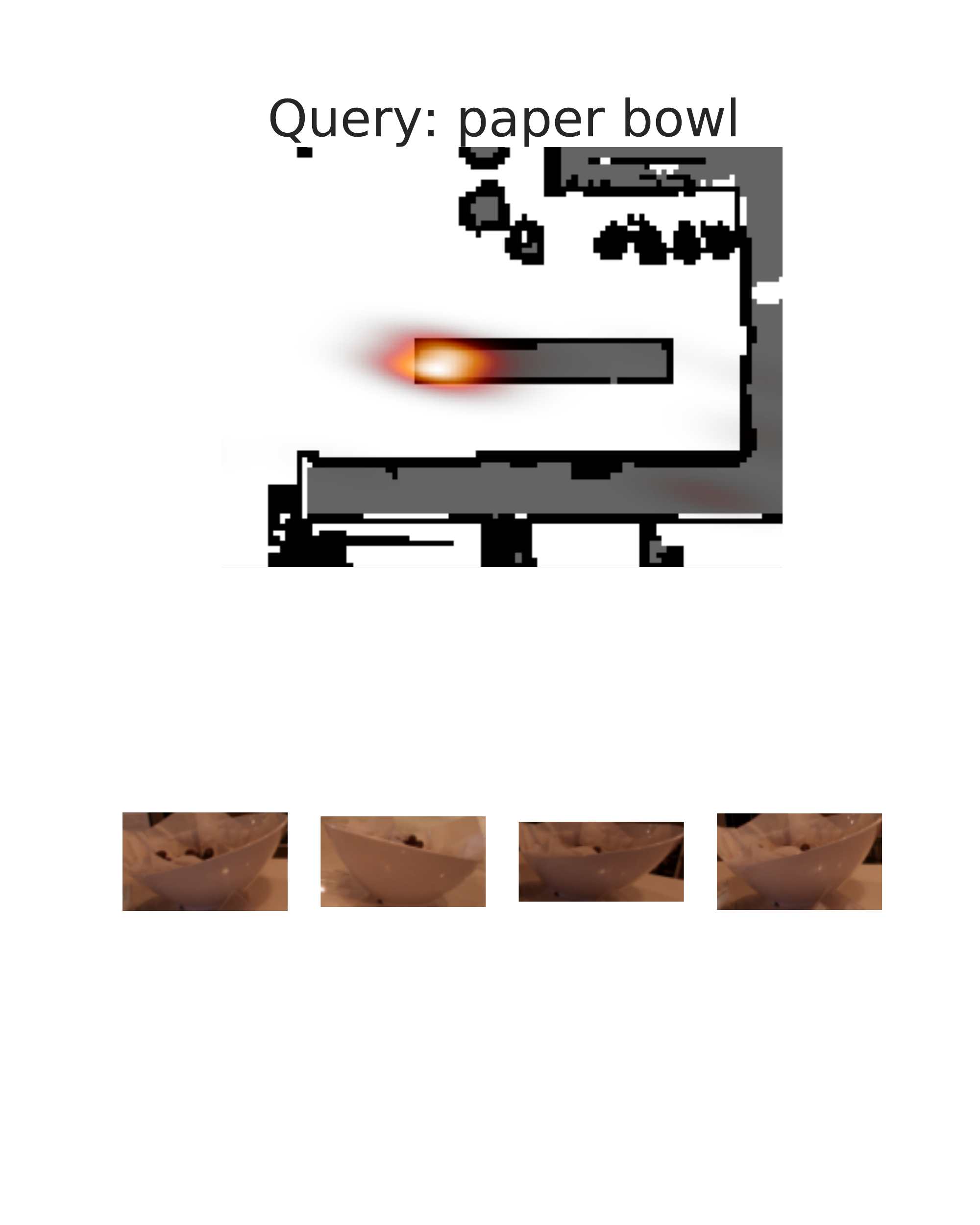}
         \caption{}
\end{subfigure}

\begin{subfigure}[b]{0.24\textwidth}
         \centering
        \includegraphics[width=\textwidth, trim={2cm 2cm 1cm 1cm},clip]{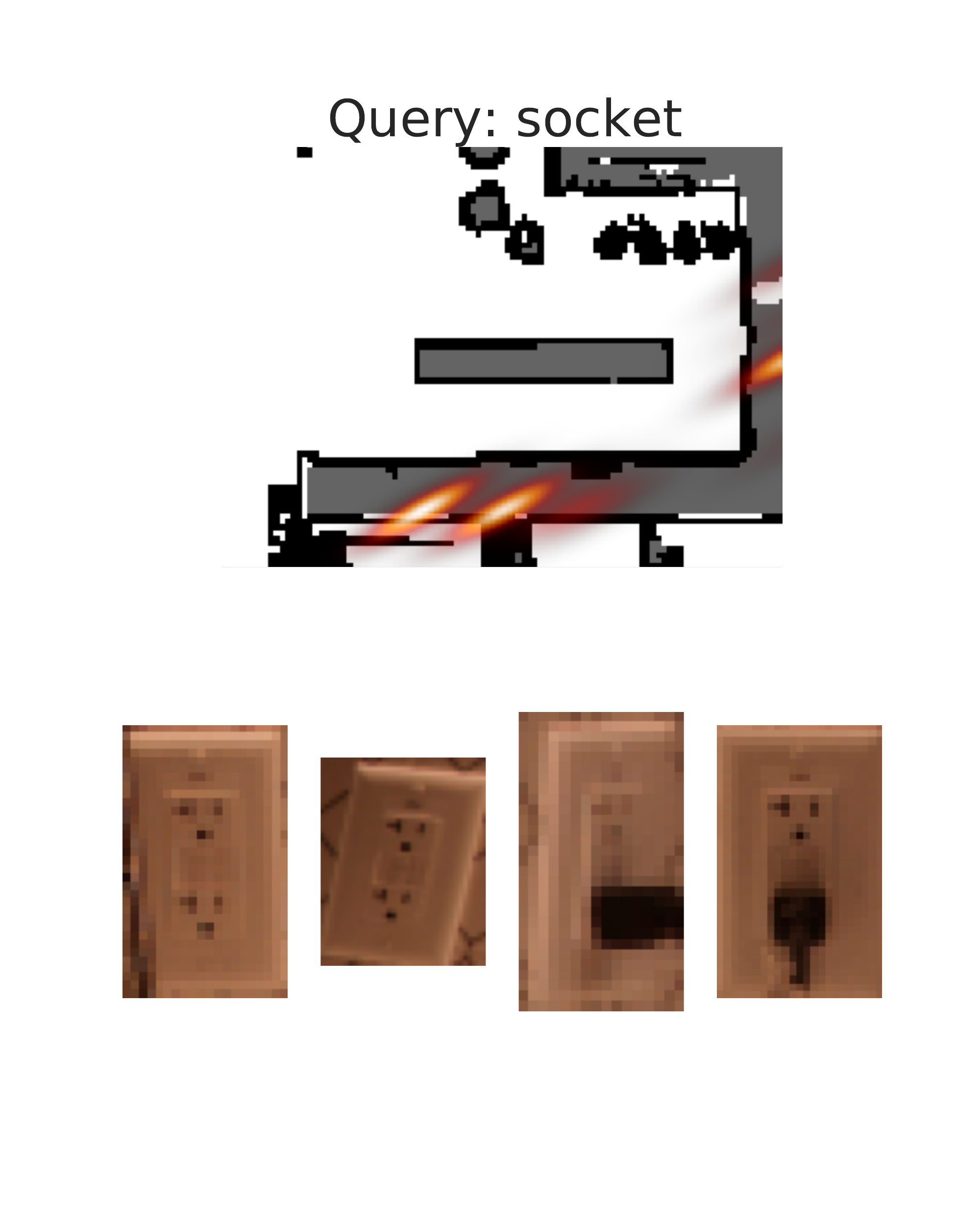}
         \caption{}
\end{subfigure}
\begin{subfigure}[b]{0.24\textwidth}
         \centering
        \includegraphics[width=\textwidth, trim={2cm 2cm 1cm 1cm},clip]{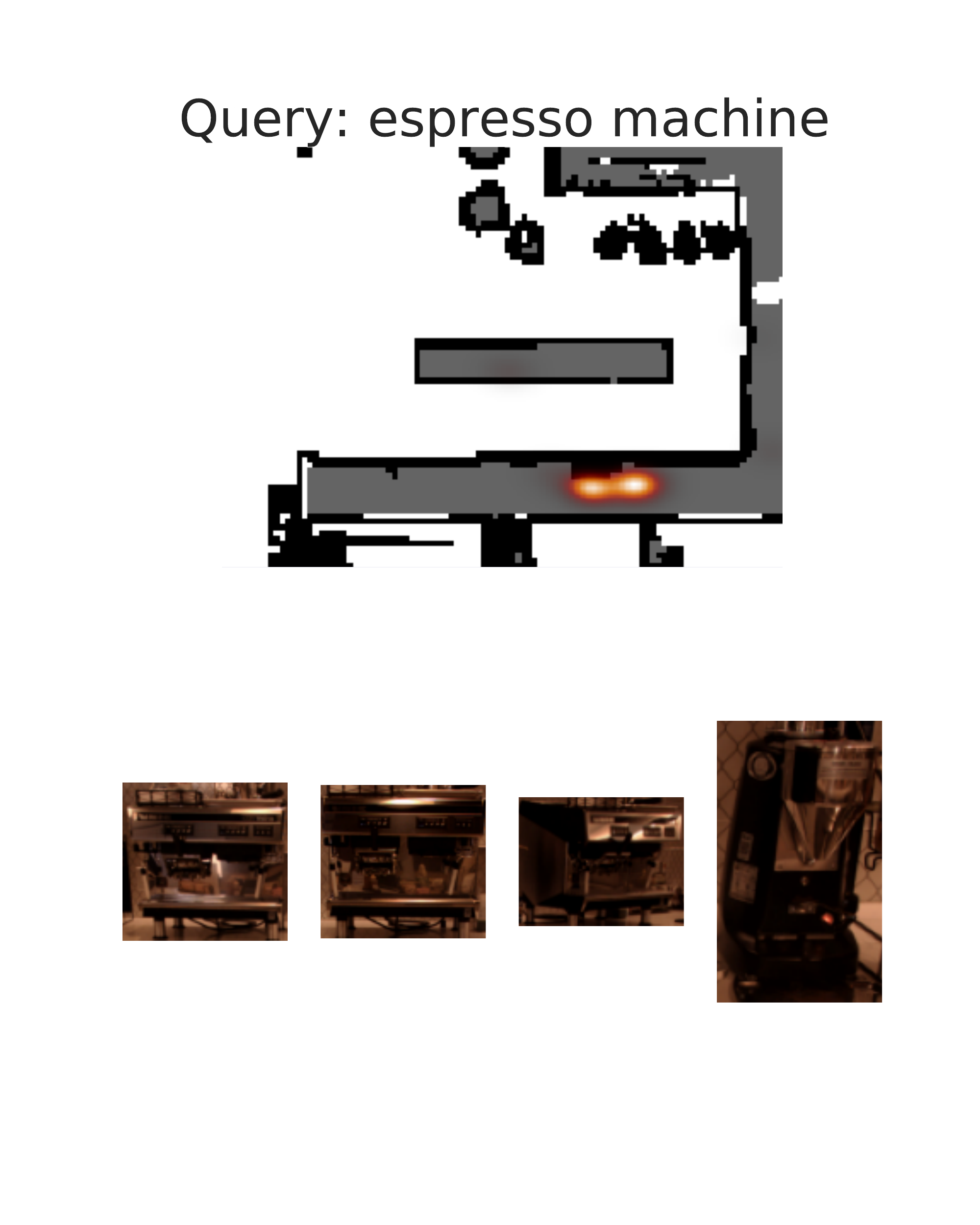}
         \caption{}
\end{subfigure}
\begin{subfigure}[b]{0.24\textwidth}
         \centering
        \includegraphics[width=\textwidth, trim={2cm 2cm 1cm 1cm},clip]{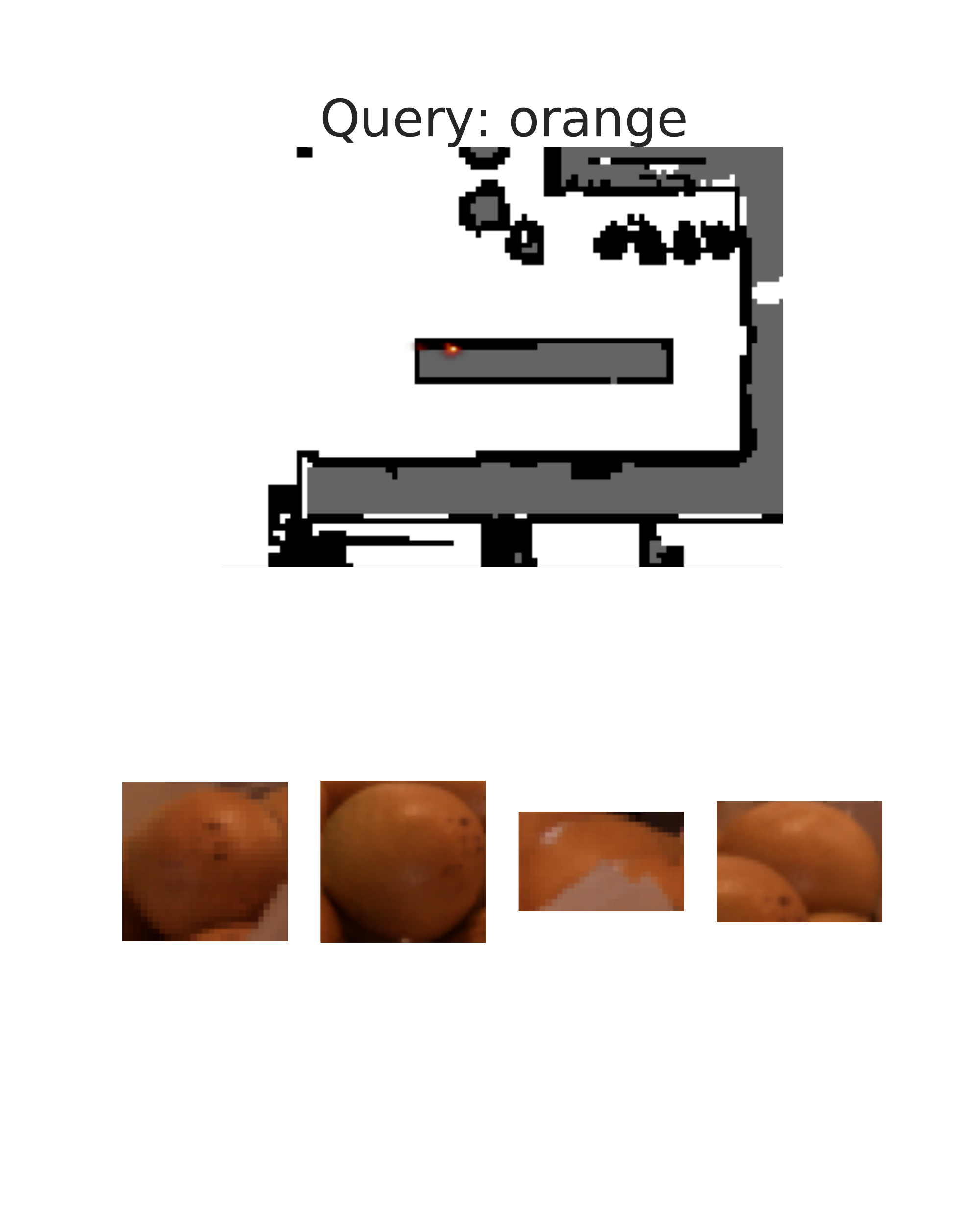}
         \caption{}
\end{subfigure}
\begin{subfigure}[b]{0.24\textwidth}
         \centering
        \includegraphics[width=\textwidth, trim={2cm 2cm 1cm 1cm},clip]{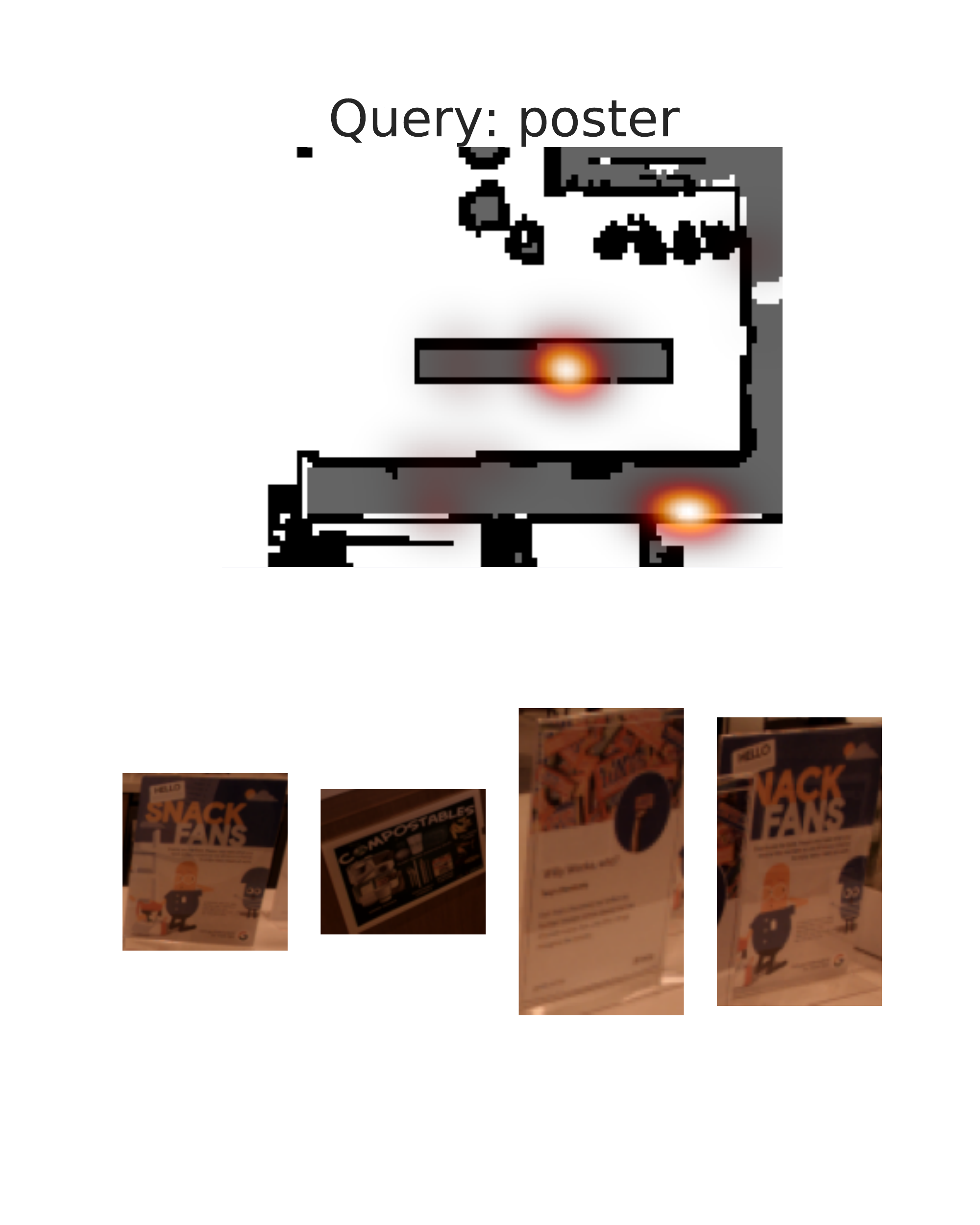}
         \caption{}
\end{subfigure}

\caption{\textbf{More examples of \algname queries.} Similar to the visualization in Fig.~\ref{fig:map}, We show the query word as title of each plot, and the heatmap of matches overlayed on the map. The top 4 RoI matches are shown below the map.\label{fig:add3}}
\end{figure*}

We show additional qualitative experiment results in Fig.~\ref{fig:add1}, Fig.~\ref{fig:add2} and Fig.~\ref{fig:add3}.

\end{document}

%% file: preamble.tex
\usepackage{url} %
\usepackage{wrapfig}

\usepackage{hyperref}
\usepackage{graphicx}
\usepackage{amsmath}
\usepackage{bm}
\usepackage{bbm}
\usepackage{mathrsfs}
\usepackage{mathtools}

\let\labelindent\relax
\usepackage{enumitem}
\usepackage{cite}

\usepackage{chngcntr}
\usepackage{apptools}
\AtAppendix{\counterwithin{prop}{section}}
\AtAppendix{\counterwithin{lem}{section}}
\AtAppendix{\counterwithin{thm}{section}}

\usepackage[font={small}]{caption}
\usepackage{subcaption}

\usepackage{color}

\usepackage{cleveref} %

\usepackage{xspace}

\usepackage{cite} %

\usepackage{amssymb}%

\usepackage{algorithm}
\makeatletter
\renewcommand*{\ALG@name}{Alg.}
\makeatother
\usepackage[noend]{algpseudocode}

\newcommand\mydots{\hbox to 1em{.\hss.\hss.}}

\usepackage{multirow}

%% file: prompt.tex
\begin{minipage}{\linewidth}

\begin{lstlisting}[basicstyle=\ttfamily\scriptsize, backgroundcolor = \color{lightgray}, keywords = {Human, Robot},  keywordstyle = {\textbf}, caption={Object proposal prompt in \algname + SayCan.}, label={lst:prompt}]
The task 'hold the snickers' may involve the following objects:snickers.
The task 'wipe the table' may involve the following objects:table, napkin, sponge, towel.
The task 'put a water bottle and an oatmeal next to the microwave' may involve the following objects:water bottle, oatmeal, microwave.
The task 'place the mug in the cardboard box' may involve the following objects:mug, cardboard box.
The task 'go to the fridge' may involve the following objects:fridge.
The task 'put a grapefruit from the table into the bowl' may involve the following objects:grapefruit, table, bowl.
The task 'can you open the glass jar' may involve the following objects:glass jar.
The task 'heat up the taco and bring it to me' may involve the following objects:taco, human, microwave oven, fridge.
The task 'hold the fancy plate with flower pattern' may involve the following objects:fancy plate with flower pattern.
The task 'put the fruits in the fridge' may involve the following objects:fridge, apple, orange, banana, peach, grape, blueberry.
The task 'get a sponge from the counter and put it in the sink' may involve the following objects:sponge, counter, sink.
The task 'empty the water bottle' may involve the following objects:water bottle, sink.
The task 'i am hungry, give me something to eat' may involve the following objects:human, candy, snickers, chips, apple, banana, orange.
The task 'go to the trash can for bottles' may involve the following objects:trash can for bottles.
The task 'put the apple in the basket and close the door' may involve the following objects:apple, basket, door.
The task 'help me make a cup of coffee' may involve the following objects:cup, coffee, mug, coffee machine.
The task 'check what time is it now' may involve the following objects:clock, watch.
The task 'let go of the banana' may involve the following objects:banana, trash can.
The task 'put the grapes in the bowl and then move the cheese to the table' may involve the following objects:grape, bowl, cheese.
The task 'find a coffee machine' may involve the following objects:coffee machine.
The task 'clean up the spilled coke' may involve the following objects:spilled coke, towel, mop, napkin, sponge.
The task 'bring me some soft drinks' may involve the following objects:human, pepsi, coke, sprite, fanta, 7 up.
The task 'boil some water' may involve the following objects:water, kettle, sink, tap.
The task 'wash the dishes' may involve the following objects:sink, tap, mug, plate, bowl, fork, spoon, knife.
The task 'place a knife and a banana to the table' may involve the following objects:knife, banana, table.
\end{lstlisting}

\end{minipage}

\begin{lstlisting}[basicstyle=\ttfamily\scriptsize, backgroundcolor = \color{lightgray}, keywords = {Human, Robot, Available, objects, Explanation},  keywordstyle = {\textbf}, caption={Task planning prompt in \algname + SayCan.}, label={lst:prompt2}]
Robot: Hi there, I'm a robot operating in an office kitchen.
You can ask me to do various tasks and I'll tell you the sequence of actions I would do to accomplish your task.
Human: Hold the snickers
Available objects are: snickers.
Explanation: Hold on means to pick it up. I will pick up the snickers.
Robot: 1. pick up the snickers
2. done.
Human: Put the trailmix here
Available objects are: trailmix.
Explanation: This is a single action of puting down the trailmix.
Robot: 1. put down the trailmix
2. done.
Human: Put a water bottle and an oatmeal next to the microwave
Available objects are: water bottle, oatmeal, microwave.
Explanation: I will find and move a water bottle to the microwave, then find and move an oatmeal to the microwave.
Robot: 1. find the water
2. pick up the water
3. go to microwave
4. put down the water
5. find the oatmeal
6. pick up the oatmeal
7. go to the microwave
8. put down the oatmeal
9. done.
Human: Put the mug in the cardboard box
Available objects are: .
Explanation: There is no mug nor cardboard box available so I will do nothing.
Robot: 1. done.
Human: Put a grapefruit from the table into the bowl
Available objects are: grapefruit, bowl, table.
Explanation: I will find and pick up the grapefruit, then find the bowl and put the grapefruit in it.
Robot: 1. find the grapefruit
2. pick up the grapefruit
3. go to the bowl
4. put down the grapefruit
5. done.
Human: Get a sponge from the counter and put it in the sink
Available objects are: sponge, sink.
Explanation: I will find a sponge first. Then I will pick up the sponge and put it down in the sink.
Robot: 1. find the sponge
2. pick up the sponge
3. go to the sink
2. pick up the sponge
3. go to the sink
4. put down the sponge
5. done.
Human: Move the water from the table to the counter
Available objects are: water, table, counter.
Explanation: The user has requested moving the water to the counter. I will find the water and pick it up, then put it on the counter.
Robot: 1. find the water
2. pick up the water
3. go to the counter
4. put down the water
5. done.
Human: Bring me some snacks
Available objects are: human, twix.
Explanation: The user has requested some snacks, twix is a type of snack. I will bring the user some twix.
Robot: 1. find the twix
2. pick up the twix
3. find the human
4. put down the twix
5. find the dried fruit
6. pick up the dried fruit
7. find the human
8. put down the dried fruit
9. done.
Human: Bring me an orange
Available objects are: human.
Explanation: There is no orange available so I will do nothing.
Robot: 1. done.
Human: Give me something to drink
Available objects are: human, pepsi, 7 up, milk, juice.
Explanation: The user has requested some drinks, pepsi is a type of drink. I will bring the user some pepsi.
Robot: 1. find the pepsi
2. pick up the pepsi
3. find the human
4. put down the pepsi
5. done.
Human: Put the apple in the basket
Available objects are: basket.
Explanation: There is no apple available so I will do nothing.
Robot: 1. done.
Human: Deliver a can of coke to me
Available objects are: coke can.
Explanation: The user has requested a coke, I will find a coke and bring it to the user.
Robot: 1. find the coke
2. pick up the coke
3. find the human
4. put down the coke
5. done.
Human: I finished my drink, can you throw it away and bring me a mountain dew
Available objects are: coke can, mountain dew.
Explanation: The user has requested to throw his drink (which appears to be a coke) away. I will put the coke in the trash, and then bring the user a mountain dew.
Robot: 1. find the sprite
2. pick up the sprite
3. go to the trash
4. put down the sprite
5. find the mountain dew
6. pick up the mountain dew
7. find the human
8. put down the mountain dew
9. done.
Human: Let go of the banana
Available objects are: banana.
Explanation: Let go simply means to put down. I will put down the banana.
Robot: 1. put down the banana
2. done.
Human: Put the grapes in the bowl and then move the cheese to the table
Available objects are: grape, bowl, cheese.
Explanation: This is a query with two parts. The first part requires finding grapes and then moving them to the bowl. The second part requires finding the cheese and moving it to the table. I will make sure to finish both parts of the task.
Robot: 1. find the grapes
2. pick up the grapes
3. go to the bowl
4. put down the grapes
5. pick up the cheese
6. go to the table
7. put down the cheese
8. tell you I'm done.
Human: Pick up the peanuts
Available objects are: human, peanut.
Explanation: This is a simple task of a single step, I will just pick up the peanuts.
Robot: 1. pick up the peanuts
2. done.
Human: Pick up the water bottle
Available objects are: .
Explanation: There is no watter bottle available so I will do nothing.
Robot: 1. done.
Human: Bring me the peanuts
Available objects are: peanut.
Explanation: The user has requested peanuts, I will find peanuts and bring them to the user.
Robot: 1. pick up the peanuts
2. find the human
3. put down the peanuts
4. done.
Human: Throw away a coffee cup
Available objects are: coffee cup, trash can.
Explanation: The user has requested me to throw away a coffee cup. Throwing away means putting something in the trash can. I will find a coffee cup, pick that up and then put it in the trash.
Robot: 1. find the coffee cup
2. pick up the coffee cup
3. go to the trash
4. put down the coffee cup
5. done.
Human: Place a knife and a banana to the table
Available objects are: knife, table.
Explanation: There is no banana available so I will do nothing.
Robot: 1. done.
Human: Throw away the fruits
Available objects are: apple, orange, banana, lime.
Explanation: The user has requested me to throw away the fruits. Throwing away means putting something in the trash can. Banana is a type of fruit that's available. I will find banana, pick that up and then put it in the trash.
Robot: 1. find the banana
2. pick up the banana
3. go to the trash
4. put down the banana
5. done.
\end{lstlisting}

%% file: obj_proposal_task.tex
\begin{minipage}{\linewidth}

\begin{lstlisting}[basicstyle=\ttfamily\scriptsize, backgroundcolor = \color{lightgray}, keywords = {Human, Robot},  keywordstyle = {\textbf}, caption={Object proposal task list, where robot needs to infer objects from tasks}, label={lst:obj_proposal_task}]
make lasagna
cook chicken tikka masala
make a sandwich
recycle the coke can
freeze the ice cream in the shopping bag
blend pineapples and mangos to make some smoothies
fillet the fish
find some container to serve the steak
compost the apple
water the plant
slice the sausages and put them into a bowl
microwave the to go box
give me something to brush my teeth
light up the room
season the steak
cook an egg
bake the apple pie
fill the paper cup with water
cut the paper in half
wash away the dusts on the cutting board
drain the rice
stir fry the bok choy
steam the dumplings
sharpen the knife
throw away the yogurt cup
\end{lstlisting}
\end{minipage}

\begin{minipage}{\linewidth}

\begin{lstlisting}[basicstyle=\ttfamily\scriptsize, backgroundcolor = \color{lightgray}, keywords = {Human, Robot},  keywordstyle = {\textbf}, caption={Object proposal task list, where the robot needs to understand complex human language inputs}, label={lst:obj_proposal_task2}]
I opened a pepsi earlier. bring me an open can
I spilled my coke, can you bring me a replacement
I spilled my coke, can you bring me something to clean it up
I accidentally dropped that jalapeno chips after eating it. Would you mind throwing it away
I like fruits, can you bring me something I would like
There is a close counter, a far counter, and a table. visit all the locations
There is a close counter, a trash can, and a table. visit all the locations
Redbull is my faviorite drink, can I have a one please
Would you bring me a coke can
Please, move the pepsi to the close counter
Can you move the coke can to the far counter
Would you throw away the bag of chips for me
Put an energy bar and water bottle on the table
Bring me a lime soda and a bag of chips
Can you throw away the apple and bring me a coke
Bring me a 7up can and a tea
Move an multigrain chips to the table and an apple to the far counter
Move the lime soda, the sponge, and the water bottle to the table
Bring me two sodas
Move three cokes to the trash can
Throw away two cokes
Bring me two different sodas
Bring me an apple, a coke, and water bottle
I spilled my coke on the table, throw it away and then bring me something to help clean
I just worked out, can you bring me a drink and a snack to recover
\end{lstlisting}
\end{minipage}

\begin{minipage}{\linewidth}

\begin{lstlisting}[basicstyle=\ttfamily\scriptsize, backgroundcolor = \color{lightgray}, keywords = {Human, Robot},  keywordstyle = {\textbf}, caption={Object proposal task list, where reference to objects contains fine grained descriptions}, label={lst:obj_proposal_task3}]

put the red can in the trash bin
put the brown multigrain chip bag in the woven basket
find the succulent plant
pick up the up side down mug
put put the apple on the macbook with yellow stickers
use the dyson vacuum cleaner
bring me the kosher salt
put the used towels in washing machine
move the used mug to the dish washer
place the pickled cucumbers on the shelf
find my mug with the shape of a donut
put the almonds in the almond jar
fill the zisha tea pot with water
take the slippery floor sign with you
give me my slippers that have holes on them
find the mug on the mini fridge
bring me the mint flavor gum
find some n95 masks
grab the banana with most black spots
fill the empty bottle with lemon juice
throw away the apple that's about to rot
throw away the rotting banana
take the box of organic blueberries out of the fridge
give a can of diet coke
open the drawer labelled as utensils
\end{lstlisting}
\end{minipage}

\begin{minipage}{\linewidth}

\begin{lstlisting}[basicstyle=\ttfamily\scriptsize, backgroundcolor = \color{lightgray}, keywords = {Human, Robot},  keywordstyle = {\textbf}, caption={Object proposal task list, where robot needs to infer objects from categories and decompose it to the right granularity}, label={lst:obj_proposal_task4}]
list some different types of masks in the house
find out what types of pastries are there in the kitchen
tell me what type of spices we have in the kitchen
find some appropriate storages for mugs
what are some protein rich food
check out what types of ingredients are available to cook a luxurious breakfast
bring me a bunch of flowers
find me some different types of Chinese dumplings in the freezer
give me a bunch of different flowers
put different kinds of common cheeses in the fridge
list all available vegetables in the fridge
give me some sweet snacks
give me some savory snacks
give me some first-aid items
mix all types of wines in the cabinet
\end{lstlisting}
\end{minipage}

%% file: task_list.tex
\begin{minipage}{\linewidth}

\begin{lstlisting}[basicstyle=\ttfamily\scriptsize, backgroundcolor = \color{lightgray}, keywords = {Human, Robot},  keywordstyle = {\textbf}, caption={Task List used in experiment. The scene setup is the same as in SayCan~\cite{saycan2022arxiv}.}, label={lst:task_list_1}]
put the coke can in the your gripper
let go of the coke can
come to the table
deliver the red bull to the close counter
throw away the water bottle
put the apple back on the far counter
bring me something to quench my thirst
bring me a fruit
bring me a bag of chips from close counter
pick up the 7up and bring it to me
pick up the water bottle and move it to the trash
pick up the apple and move it to the far counter
Please, move the pepsi to the close counter
Would you throw away the bag of chips for me
Redbull is my favorite drink, can I have one please
Can you throw away the apple and bring me a coke
How would you bring me an apple, a coke, and water bottle
I just worked out, can you bring me a drink and a snack to recover?
Please, move the ppsi to the close cuonter
Would you throw away the bag of chpis for me
\end{lstlisting}

\end{minipage}

\begin{minipage}{\linewidth}

\begin{lstlisting}[basicstyle=\ttfamily\scriptsize, backgroundcolor = \color{lightgray}, keywords = {Human, Robot},  keywordstyle = {\textbf}, caption={Task List used in experiment, the scene set up is an office kitchen full of objects, plus testing objects: multigrain chip, basket, plant, sink, apple, first aid station, coke, sink, rice chip bag, coffee machine, water bottle, paper cup, lime sparkling water(green), yellow sign, snack jars of nuts, snack jar of dried fruits, snack jar of gums, snack jar of candy, mug, water fountain machine, tv, tea bottle, box of tea, energy cup, paper bowl, clip board, compost bin.}, label={lst:task_list_2}]
Put the brown multigrain chip in the woven basket
Water the potted plant
Wash the apple
Move a can of soda to the first aid station
Put the red can in the sink
Put the green chip bag in front of the coffee machine
Help me put bottled water near the paper cup
Compost the apple
Show me where is the first aid station
Put the green can besides the yellow sign
Go to the snack jar of nuts
Microwave the apple
Moisterize the green plant
I want to watch TV, can you get a bottle of tea and put it there
Place the plastic bottle next to the snack jar
Put the chip bag in front of the box of tea
Move the apple to the plant and water the plant
Find a disposable cup
Please help me put the energy bar in the paper bowl
Can you help me find where my clipboard is
\end{lstlisting}

\end{minipage}

\begin{minipage}{\linewidth}

\begin{lstlisting}[basicstyle=\ttfamily\scriptsize, backgroundcolor = \color{lightgray}, keywords = {Human, Robot},  keywordstyle = {\textbf}, caption={Task List used in experiment. The scene setup is an office kitchen full of other common objects but without coke, green can, protein bar, green chip bag, bottle of tea, apple, coke can, multigrain chip.}, label={lst:task_list_3}]
Put the brown multigrain chip in the woven basket
Water the potted plant
Wash the apple
Move a can of soda to the first aid station
Put the red can in the sink
Put the green chip bag in front of the coffee machine
Help me put bottled water near the paper cup
Compost the apple
Put the green can besides the yellow sign
Microwave the apple
Moisterize the green plant
I want to watch TV, can you get a bottle of tea and put it there
Put the chip bag in front of the box of tea
Move the apple to the plant and water the plant
Please help me put the energy bar in the paper bowl
\end{lstlisting}

\end{minipage}